\pdfoutput=1
\documentclass[]{macro/neu_msthesis}
\usepackage{tabularx}
\usepackage{comment}
\usepackage{float}
\usepackage{acronym}
\usepackage{lipsum}
\usepackage{color}
\usepackage[style=numeric]{biblatex}
\DeclareSourcemap{
  \maps{
    \map{
      \pertype{article}
      \step[fieldset=language, null]
      \step[fieldset=url, null]
      \step[fieldset=issn, null]
      \step[fieldset=isbn, null]
      \step[fieldset=note, null]
      \step[fieldset=editor, null]
      \step[fieldset=urldate, null]
      \step[fieldset=file, null]
    }
  }
}
\DeclareSourcemap{
  \maps{
    \map{
      \pertype{inproceedings}
      \step[fieldset=language, null]
      \step[fieldset=url, null]
      \step[fieldset=issn, null]
      \step[fieldset=isbn, null]
      \step[fieldset=note, null]
      \step[fieldset=editor, null]
      \step[fieldset=urldate, null]
      \step[fieldset=file, null]
    }
  }
}
\DeclareSourcemap{
  \maps{
    \map{
      \pertype{incollection}
      \step[fieldset=language, null]
      \step[fieldset=url, null]
      \step[fieldset=issn, null]
      \step[fieldset=isbn, null]
      \step[fieldset=note, null]
      \step[fieldset=editor, null]
      \step[fieldset=urldate, null]
      \step[fieldset=file, null]
    }
  }
}
\usepackage[usenames,dvipsnames]{xcolor}
\usepackage{tcolorbox}
\usepackage{array}
\usepackage{colortbl}
\usepackage{booktabs}
\usepackage{rotating}
\usepackage{algpseudocode}
\usepackage{algorithm}

\title{Reinforcement Learning-Based Model Matching in COBRA, a Slithering Snake Robot}

\author{Harin Kumar Nallaguntla}

\nuid{002751978}

\dept{Electrical and Computer Engineering}

\degreename{Robotics}

\field{Robotics}

\submitdate{April 2024}

\numberofmembers{3}
\principaladviser{Dr. Alireza Ramezani}
\secondadviser{Dr. Gunar Schirner}
\firstreader{Dr. Michael Everett}
\usepackage{amsfonts,amssymb,amsmath}			% special math characters
\usepackage{times}		% use Postscript fonts
\usepackage{bm}

\newcommand{\ifno}[1]{}

\usepackage{multirow}
\makeindex
\usepackage{makeidx}
\usepackage{acronym}

\usepackage{url}
\usepackage{epstopdf}
\usepackage[]{graphicx}

\ifnum\pdfoutput>0
\usepackage[pdftex,%                    % hyper-references for ps2pdf
bookmarks=true,%                        % generate bookmarks ...
bookmarksnumbered=true,%                % ... with numbers
hypertexnames=false,%                   % needed for correct links to figures
breaklinks=true           %Allows link text to break across lines;
]{hyperref}
\else
\usepackage[hypertex,%                  % hyper-references for ps2pdf
bookmarks=true,%                        % generate bookmarks ...
bookmarksnumbered=true,%                % ... with numbers
hypertexnames=false,%                   % needed for correct links to figures
breaklinks=true           %Allows link text to break across lines;
]{hyperref}
\fi

\hypersetup{				% setup PDF information fields
pdfauthor = {\authorRef},
pdftitle = {\titleRef},
pdfsubject = {\expandafter{\degreeRef} thesis submitted to Northeastern University},
pdfkeywords = {add keywords here}
pdfcreator = {LaTeX with hyperref package},
pdfproducer = {dvips + ps2pdf}}

\usepackage{microtype}

\begin{document}

% add a pdf bookmark to the cover page
\pdfbookmark[1]{Cover}{cover}

% --- title page ---
\titlepage

% --- front matter ---
\begin{frontmatter}
% %%%%%% NO SIGNATURE PAGE WILL BE GENERATED, PLEASE DOWNLOAD FROM COE WEB SITE (READ file Readme.pdf in this folder!)
%\signaturepage
% dedication

% dedication.tex:

\begin{dedication}
To my mom.
\end{dedication}

% table of content (add bookmark for convenience)

\pdfbookmark[1]{Table of Contents}{contents}
\tableofcontents
\listoffigures
\newpage\ssp
\listoftables

% include a list of Acronyms (comment out if no acronyms are specified)
% acronyms.tex

\chapter*{List of Acronyms}
% \addcontentsline{toc}{chapter}{List of Acronyms}

% below is the list of acronym definitions, place them in alphabetical order
% since they will not be sorted again. 
\begin{acronym}
\acro{ROS}{Robot Operating System}.

\acro{ODE}{Open Dynamics Engine}.

\acro{AI}{Artificial Intelligence}.

\acro{DRL}{Deep Reinforcement Learning}.

\acro{GPU}{Graphics Processing Unit}.

\acro{PPO}{Proximal Policy Optimization}.

\acro{CPG}{Central Pattern Generators}.

\acro{CFM}{Constraint Force Mixing}.

\acro{ERP}{Error Reduction Parameter}.

\acro{SDR}{Structured Domain Randomization}.

\acro{COBRA}{Crater Observing Bio-inspired Rolling Articulator}.

\acro{SAC}{Soft Actor Critic}.

\acro{MDP}{Markov Decision Process}.

\acro{DQN}{Deep Q Network}.

% \acro{MPC}{Model Predictive Control}.

% \acro{GRF}{Ground Reaction Forces}.

% \acro{HZD}{Hybrid Zero Dynamics}.

% \acro{ZMP}{Zero Moment Point}.

% \acro{CoM}{Center of Mass}.

% \acro{CoP}{Center of Pressure}.

% \acro{HF}{Hip Frontal}.

% \acro{HS}{Hip Sagittal}.

\end{acronym}

%% include any of the front matter files that contain text
%% attention the input does cause a page break, the include on
%% the other hand does not
% acknowledgements.tex:

\begin{acknowledgements}

I am deeply thankful to Dr. Alireza Ramezani, my advisor, for his unwavering support and guidance throughout the completion of this thesis. I would also like to extend my gratitude to Dr. Gunar Schirner, my co-advisor, whose invaluable insights were crucial in shaping this work. I also express my appreciation to my fellow lab mates, including Adarsh, Kruthika, Aditya, and others, for their invaluable assistance in conducting physical testing, handling hardware components, and contributing to the formulation of solutions to challenges. Finally, I express my gratitude for my family, whose unwavering belief in me has been the primary source of motivation throughout my journey marked by diligence and commitment.

\end{acknowledgements}
% abstract.tex:

\begin{abstract}
{
This work employs a reinforcement learning-based model identification method aimed at enhancing the accuracy of the dynamics for our snake robot, called COBRA. Leveraging gradient information and iterative optimization, the proposed approach refines the parameters of COBRA's dynamical model such as coefficient of friction and actuator parameters using experimental and simulated data. Experimental validation on the hardware platform demonstrates the efficacy of the proposed approach, highlighting its potential to address sim-to-real gap in robot implementation.
}

\end{abstract}

\end{frontmatter}

% --- body of the document ---
\pagestyle{headings}

%% include each chapter like below
% intro.tex:

\chapter{Introduction}
\label{chap:intro}

Developing effective locomotion controllers for mobile robots is a challenging task. Recent advancements in \ac{DRL} have shown a promising path in automating the design of robotic locomotion controllers \cite{xie_feedback_2018} \cite{radosavovic_learning_2023} \cite{lee_learning_2020} \cite{liu_learning_2020}. However, the amount of training data required by most \ac{DRL} methods is often impractical to obtain in high-risk tasks such as manipulation, locomotion or loco-manipulation tasks. While computer simulation offers a safe and efficient environment for learning motor skills, policies trained solely in simulation often struggle to transfer seamlessly to real hardware due to various modeling discrepancies between the simulated and real environments, commonly referred to as the "Simulation-to-Reality Gap" \cite{yu_sim--real_2019} \cite{qin_sim--real_2019} \cite{escontrela_zero-shot_2020}.

This problem arises due to several factors, including but not limited to differences in frictional coefficients, inaccuracies in joint dynamics modeling \cite{noauthor_open_nodate}, and variations in sensor noise levels between simulation environments and real-world scenarios. The simulation-to-reality discrepancies hinder the seamless transition of control policies and locomotion strategies from simulation experiments to practical applications \cite{yu_sim--real_2019}, leading to suboptimal performance and limited generalization capabilities of robotic systems \cite{liu_learning_2020}.

Let S represent the simulated environment, $R$ denote the real-world environment, and 
$P(S)$ and $P(R)$ represent the performance metrics of a given robotic system in simulation and reality, respectively \cite{weng_domain_2019} \cite{vuong_how_2019}. The goal is to minimize the difference or discrepancy between $P(S)$ and $P(R)$ to achieve effective transferability of control policies and behaviors from simulation to reality. Then, Mathematically, the Sim-to-Real problem can be formulated as an optimization task:
\begin{equation}
    \theta_{\text{min}} = \arg\min_{\theta} \ D(P(S,\theta), P(R))
\end{equation}
where, $\theta$ represents the dynamic parameters that represent the simulation. $\theta_{\text{min}}$ represents the minimum value of the parameter theta ($\theta$) that minimizes the discrepancy. $D(P(S,\theta), P(R))$ is the discrepancy measure between the simulated performance $(P(S,\theta))$ and the real-world performance $(P(R))$.

To address the challenges posed by the Simulation-to-Reality Gap, recent research \cite{cubuk_autoaugment_2019} \cite{ruiz_learning_2019} has explored more sophisticated system identification procedures to enhance model accuracy \cite{yu_sim--real_2019}, training the policy on real-robot while improving accuracy of the simulator (domain adaption) \cite{chebotar_closing_2019}, and developed robust control policies capable of generalizing across diverse simulated environments (domain randomization) \cite{lee_learning_2020} \cite{hoeller_anymal_2023}. Although these approaches have demonstrated decent transferability of learned policy from simulator to real robot, they require training for atleast 1000 hrs in simulator and/or hundreds of rollouts on the real robot \cite{chebotar_closing_2019}.

\begin{figure}
  \centering
  \includegraphics[width=0.8\linewidth]{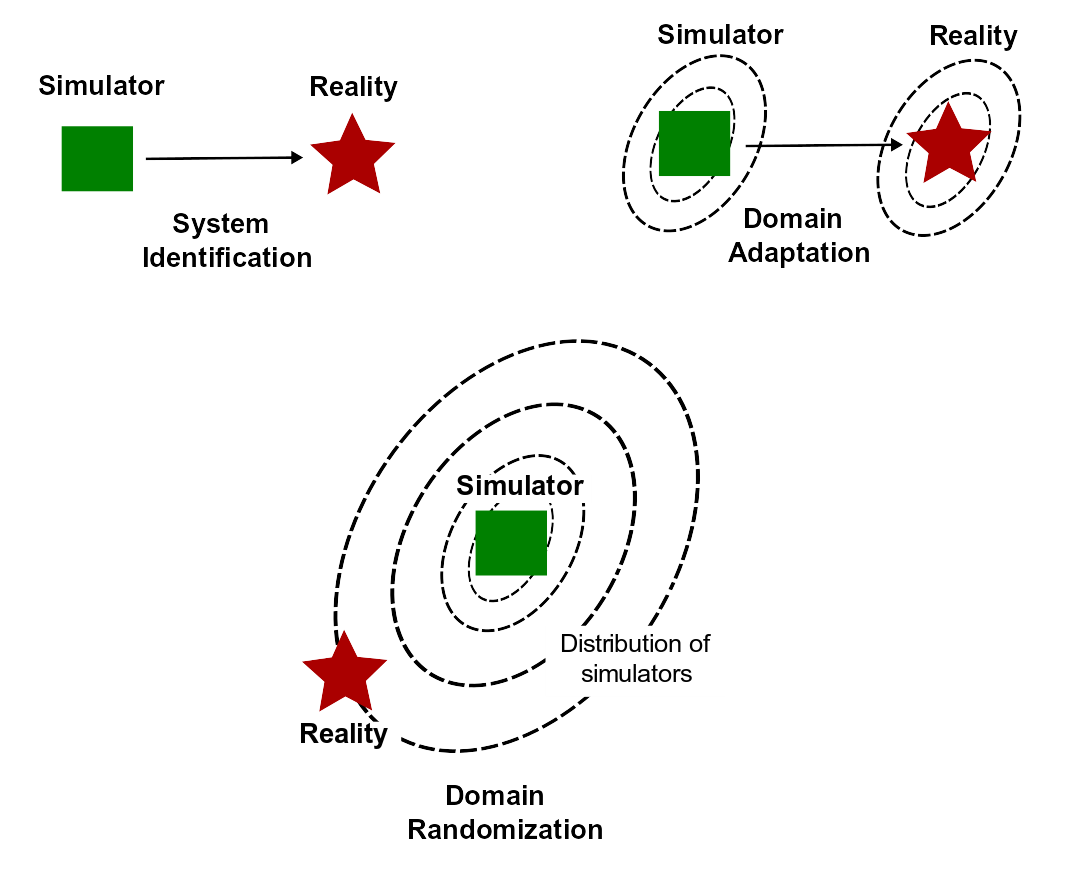} 
  \caption{Popular Solutions for Solving Sim2real Problem} \cite{weng_domain_2019}
  \label{fig:example}
\end{figure}

\section{State-Of-The-Art Modeling Tools}

In this section, we briefly describe state-of-the-art modeling tools applied successfully on multi-joint robots. Simulators play a crucial role in the development and testing of \ac{DRL}-based gait design for robots. They provide a virtual environment where researchers and engineers can simulate various scenarios, train control policies, and evaluate robot behavior without the need for physical hardware. In the context of \ac{DRL}-based locomotion strategies, simulators such as Webots \cite{noauthor_cyberbotics_nodate}, Simscape \cite{noauthor_httpswwwmathworkscomproductssimscapehtml_nodate}, Gazebo \cite{noauthor_gazebo_nodate}, Pybullet \cite{noauthor_solvers_nodate}, MuJoCo \cite{noauthor_mujoco_nodate} and Isaac Sim \cite{noauthor_isaac_nodate} offer different features, capabilities, and trade-offs.

\textbf{1. Webots:} It is a popular commercial simulator widely used for robotics research and development. It offers a user-friendly interface, a diverse library of robot models, and support for various programming languages, including Python and C++. Webots \cite{noauthor_cyberbotics_nodate} excels in providing decent physics simulation, realistic sensor models, and visualization tools, making it suitable for training RL-based locomotion policies. While Webots offers a user-friendly interface and decent physics simulation, it relies on the \ac{ODE} \cite{noauthor_open_nodate}, which has limitations in approximating friction and providing robust support for joint damping. These drawbacks can affect the accuracy and realism of simulations, especially for complex robotic systems requiring precise friction modeling and joint behaviors.

\textbf{2. Simscape: }It is a part of the MATLAB/Simulink environment, is another simulation platform used for modeling and simulating physical systems, including robotic mechanisms. Simscape offers a modular approach to system modeling, allowing users to design complex robotic systems with interconnected components and actuators. It integrates seamlessly with MATLAB for control algorithm development, data analysis, and visualization. Simscape \cite{noauthor_httpswwwmathworkscomproductssimscapehtml_nodate} utilizes the ODE4 equation and smooth stick-slip friction with a spring-damper contact model, providing good fidelity in physical modeling. However, it is incompatible with training RL agents directly within the simulation environment. Integrating RL algorithms and custom locomotion strategies into Simscape simulations may require additional effort due to its emphasis on detailed physics modeling over RL-specific functionalities.

\textbf{3. Gazebo: }It is an open-source simulator widely adopted in the robotics community, especially for \ac{ROS} integration \cite{noauthor_gazebo_nodate}. It provides a highly customizable and extensible framework for simulating robotic platforms, sensors, and environments. Gazebo's support for ROS enables seamless integration with RL libraries such as OpenAI Gym and PyBullet, facilitating the development of RL-based locomotion strategies. Additionally, Gazebo offers plugins for sensor simulation, dynamic control, and physics tuning, making it suitable for realistic and scalable robotic simulations. However, it also uses the Open Dynamics Engine (ODE) \cite{noauthor_open_nodate}, which may not provide optimal solutions for friction modeling and joint damping. Users may encounter challenges in achieving high-fidelity simulations, particularly when simulating intricate contact dynamics and damping effects.

\textbf{4. PyBullet:} PyBullet is an open-source physics engine and simulation environment designed for robotic simulations, physics-based games, and machine learning research \cite{noauthor_solvers_nodate}. It provides a Python interface to the Bullet Physics SDK, offering fast and accurate physics simulation for a wide range of robotic systems. PyBullet's integration with Python and popular machine learning libraries such as TensorFlow and PyTorch makes it suitable for developing RL-based locomotion strategies. However, PyBullet may have limitations in terms of sensor modeling and visualization compared to other simulators and it utilizes the Bullet Physics Engine, which may have limitations in accurately simulating scenarios with detailed friction modeling and joint damping. Users may face challenges in configuring the solver for optimal performance in scenarios requiring precise control over friction forces and joint behaviors.

\textbf{5. MuJoCo:} MuJoCo (Multi-Joint dynamics with Contact) \cite{noauthor_mujoco_nodate} is a commercial physics engine known for its efficient and stable simulation of articulated rigid bodies and contact dynamics. It is commonly used in robotics research and reinforcement learning applications due to its fast simulation speed and accurate physics modeling. MuJoCo provides a high degree of customization for designing robotic systems and environments, making it suitable for simulating complex locomotion behaviors. However, its integration with \ac{DRL} frameworks may require additional development. Additionally, MuJoCo's physics modeling (using \ac{ODE}), including friction approximation and joints modeling, may not always meet the requirements of complex robotic simulations, impacting the fidelity of \ac{DRL}-based locomotion strategies.

\textbf{6. Isaac Sim:} Isaac Sim is a simulation platform developed by NVIDIA, specifically designed for robotics and \ac{AI} research \cite{noauthor_isaac_nodate}. It offers a comprehensive set of tools and libraries for simulating robotic systems, sensor models, and AI algorithms. Isaac Sim leverages NVIDIA's \ac{GPU}-accelerated physics simulation for fast and realistic simulations of complex robotic behaviors. Its integration with NVIDIA's Isaac SDK and Isaac Gym framework enables seamless development and testing of \ac{DRL}-based locomotion strategies. Isaac Sim's LULA solver offers a good friction model and realistic simulation of complex robotic behaviors. However, its reliance on NVIDIA's \ac{GPU}-accelerated physics simulation may pose challenges for researchers lacking access to compatible hardware with sufficient GPU compute power. This limitation can hinder the training and testing of RL-based locomotion strategies within Isaac Sim, especially for resource-constrained environments.

\section{Sim-to-Real Transfer for RL-based Strategies}

Sim-to-Real transfer in the context of RL-based locomotion strategies represents a critical frontier in robotics research, aiming to bridge the gap between simulated environments and real-world deployment \cite{kolvenbach_traversing_2021} \cite{frey_locomotion_2022}. The ability to develop locomotion controllers in simulation and transfer them seamlessly to physical robots is of paramount importance for practical applications in challenging terrains and dynamic environments. This section briefly delves into the advancements in achieving effective Sim-to-Real transfer for RL-based locomotion strategies.

\begin{figure}
  \centering
  \includegraphics[width=1.0\linewidth]{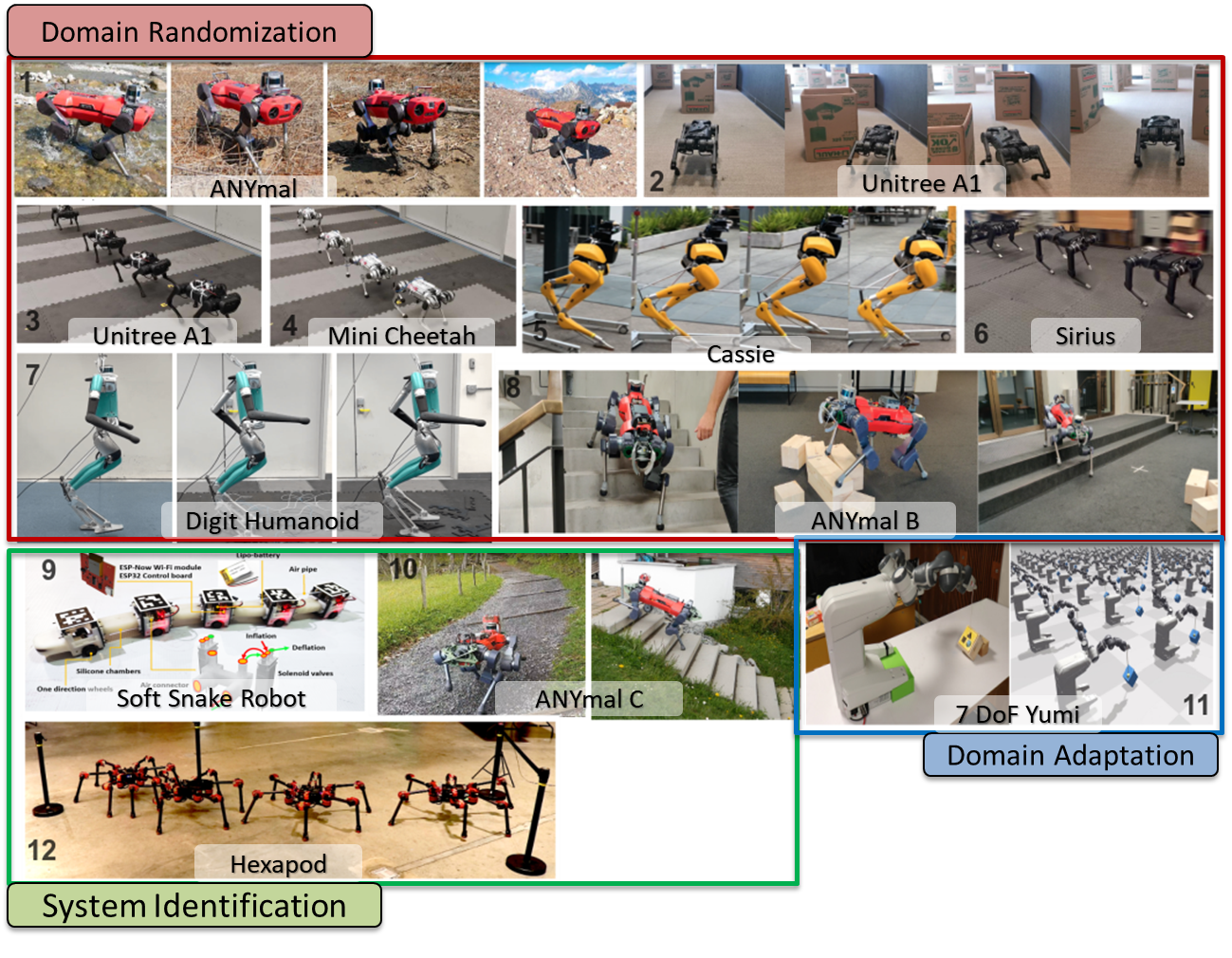} 
  \caption{Shows successful transfer of an RL-based policy on 1) ANYmal robot \cite{lee_learning_2020} 2) \& 6) Unitree A1 \cite{yang_learning_2022} \cite{feng_genloco_2022} 3) Hexapod \cite{li_learning_2020} 4) Cassie \cite{li_reinforcement_2021} 5) Sirius \cite{feng_genloco_2022} 7) Mini Cheetah \cite{feng_genloco_2022} 8) 7-DoF Yumi Robot \cite{chebotar_closing_2019} 9) \& 13) ANYmal C \cite{miki_learning_2022} 10) ANYmal B \cite{rudin_learning_2022} 11) Digit humanoid \cite{radosavovic_learning_2023} 12) Soft Snake Robot \cite{liu_reinforcement_2023}}

   \label{fig:sota}
\end{figure}

Several notable approaches have been developed by different teams to address the challenge of training RL-based locomotion policies in simulation and transferring them to physical robots. In the paper \cite{lee_learning_2020} introduces a comprehensive framework that integrates a privileged student-teacher training setup and domain randomization to tackle the sim2real problem effectively. This approach enables the training of locomotion policies in a simulator and their successful transfer to real robots by dynamically adjusting simulation parameters to match real-world conditions.

Another significant contribution comes from the paper \cite{yu_sim--real_2019}, which emphasizes the importance of pre-sysID and post-sysID stages in system identification for simulation. By collecting real-world trajectories and using Bayesian Optimization, their method approximates accurate parameters for simulation, guiding subsequent policy learning for locomotion tasks. This approach enhances the adaptability and effectiveness of trained policies by ensuring that the simulation closely reflects real-world scenarios, thus improving the transferability of policies from simulation to physical robots.

Additionally, Qin et al. \cite{qin_sim--real_2019} focuses on gait planning for six-legged robots using \ac{DRL} and \ac{PPO}. Their approach leverages an Actor-Critic network to train the robot's gait adaptively, incorporating curriculum learning and simulations in a simplified environment. This methodology accelerates the training process and demonstrates the reliability of DRL-based locomotion strategies for navigating challenging terrains, showcasing the potential for practical applications in real-world scenarios with six-legged robots.

Sim-to-Real transfer is a crucial research area essential for implementing RL-based locomotion strategies effectively in real-world robotic systems. Researchers are actively developing innovative methodologies and leveraging technological advancements to bridge the Reality Gap and realize the full potential of RL in dynamic and challenging terrains. However, it's important to note that many existing methods for Sim-to-Real transfer often demand significant simulator training time, utilizing computationally intensive techniques like domain randomization, or necessitate the collection of extensive real robot data. There exists a pressing need for a comprehensive framework that can overcome these challenges. Such a framework would not rely heavily on domain randomization during simulator training and could build a realistic simulator with only a small amount of real robot data, thereby streamlining the Sim-to-Real transfer process and enhancing the scalability and applicability of RL-based locomotion strategies in real-world scenarios.

\section{Modeling Frameworks Background}

Drawing inspiration from Silicon Synapse Lab's modeling frameworks, I have delved into the sim2real problem and the intricate interactions of COBRA \cite{salagame_how_2023} with its environment. Within the lab's innovative ecosystem, various bio-inspired robots \cite{salagame_non-impulsive_2024} \cite{ramezani_aerobat_2022} \cite{salagame_letter_2022} \cite{dangol_control_2021} have been developed, each paired with specific modeling frameworks \cite{sihite_optimization-free_2021} designed to address unique control challenges. Below, I provide a succinct overview of these robots and the tailored modeling approaches crafted at Silicon Synapse Lab.

Aerobot \cite{sihite_integrated_2021} \cite{ramezani_biomimetic_2017} \cite{hoff_reducing_2017} \cite{ramezani_describing_2017}, an innovative bioinspired flapping drone known as Aerobat, embodies a groundbreaking actuation framework designed to emulate the intricate flight capabilities of flying animals such as bats \cite{sihite_computational_2020} \cite{ramezani_towards_2020} \cite{hoff_trajectory_2019} \cite{hoff_synergistic_2016}. Unlike conventional flapping wing designs, Aerobot boasts dynamically versatile wing conformations and 14 body joints, omitting a tail for enhanced maneuverability. This design presents significant challenges in achieving closed-loop feedback and precise actuation \cite{sihite_enforcing_2020} \cite{gillula_design_2010} \cite{ramezani_nonlinear_nodate} \cite{ramezani_lagrangian_2015}. However, by integrating mechanical intelligence and control, the framework introduces small yet impactful adjustments through low-power actuators called primers, harnessing the computational structures of the robot \cite{sihite_wake-based_2022} \cite{sihite_unsteady_2022} \cite{ramezani_bat_nodate} \cite{mandralis_minimum_2023}. This approach enables precise regulation of joint motion, contributing to Aerobat's untethered flight capabilities and paving the way for advancements in biomimetic aerial locomotion \cite{noauthor_integrated_nodate} \cite{sihite_bang-bang_2022} \cite{syed_rousettus_2017} \cite{di_luca_bioinspired_2017} \cite{lessieur_mechanical_2021}.

The Husky robot \cite{salagame_letter_2022} embodies a cutting-edge design inspired by nature's multi-modal locomotion strategies, blending legged and aerial mobility to navigate diverse terrains with remarkable flexibility \cite{sihite_efficient_2022} \cite{krishnamurthy_towards_nodate}. Developed at Northeastern University, this quadrupedal marvel integrates thrusters for aerial stability during dynamic maneuvers, overcoming challenges typical of aerial-legged systems. Utilizing a polynomial approximation of its dynamics known as HROM (Husky Reduced Model) \cite{manjikian_towards_2022}, the robot employs an optimal control framework to tackle narrow paths akin to natural animal navigation \cite{ramezani_generative_2021} \cite{salagame_quadrupedal_2023} \cite{sihite_optimization-free_2021} \cite{dangol_hzd-based_2021}. Experimental validation and simulations in high-fidelity Simscape environments showcase the Husky's prowess in executing open-loop walking gaits, paving the way for further advancements in this versatile robotic platform \cite{sihite_unilateral_2021} \cite{buss_preliminary_2014}.

Harpy \cite{dangol_control_2021}, a groundbreaking thruster-assisted bipedal robot developed at Northeastern University, is designed to tackle the formidable challenges inherent in dynamic locomotion for bipedal walkers \cite{dangol_performance_2020} \cite{dangol_feedback_2020} \cite{sihite_dynamic_2023} \cite{kim_bipedal_2021}. The complexity arises from factors like underactuation, susceptibility to external disturbances, and the computational demands of full-dynamics and environmental interactions \cite{park_finite-state_2013}. Harpy's innovative design integrates thrusters strategically placed on its torso, alongside eight actuators distributed across its legs. This configuration enables not only frontal dynamics stabilization and fast constraint satisfaction but also facilitates agile maneuvers such as dodging fallovers and navigating obstacles through controlled jumps \cite{sihite_unilateral_2021} \cite{liang_rough-terrain_2021} \cite{dangol_reduced-order-model-based_2021} \cite{de_oliveira_thruster-assisted_2020}. The research \cite{pitroda_dynamic_2023} focuses on optimizing feedback mechanisms for Harpy's thruster-assisted walking, ensuring stability, hybrid invariance, and robustness in gait execution, thus pushing the boundaries of bipedal robotics towards enhanced locomotion capabilities \cite{dangol_performance_2020} \cite{dangol_towards_2020} \cite{buss_preliminary_2014}.

M4 \cite{sihite_multi-modal_2023} \cite{sihite_demonstrating_2023}, short for Multi-Modal Mobility Morphobot, represents a leap in robotic design inspired by nature's resilient locomotion strategies. Drawing from animals like Chukars and Hoatzins, which adeptly repurpose appendages for diverse movements, M4 embodies unparalleled versatility in navigating complex terrains \cite{mandralis_minimum_2023} \cite{sihite_efficient_2022}. This robot seamlessly transitions between modes such as flying, rolling, crawling, and balancing, utilizing its multi-functional components to overcome obstacles and explore unstructured environments with autonomy. The research \cite{rajput_towards_2023} focuses on enhancing M4's autonomy through advanced perception-navigation pipelines, deep learning-based mode transitions, and energy-efficient path planning. By showcasing intelligent mode switching, efficient navigation, and reduced energy consumption, this work lays the foundation for fully autonomous multi-modal robots, setting the stage for their deployment in exploration and rescue missions, and marking a significant stride in the evolution of intelligent and versatile robotic systems.

\section{Objectives and Outline of Thesis}

\begin{figure}
  \centering
  \includegraphics[width=0.6\linewidth]{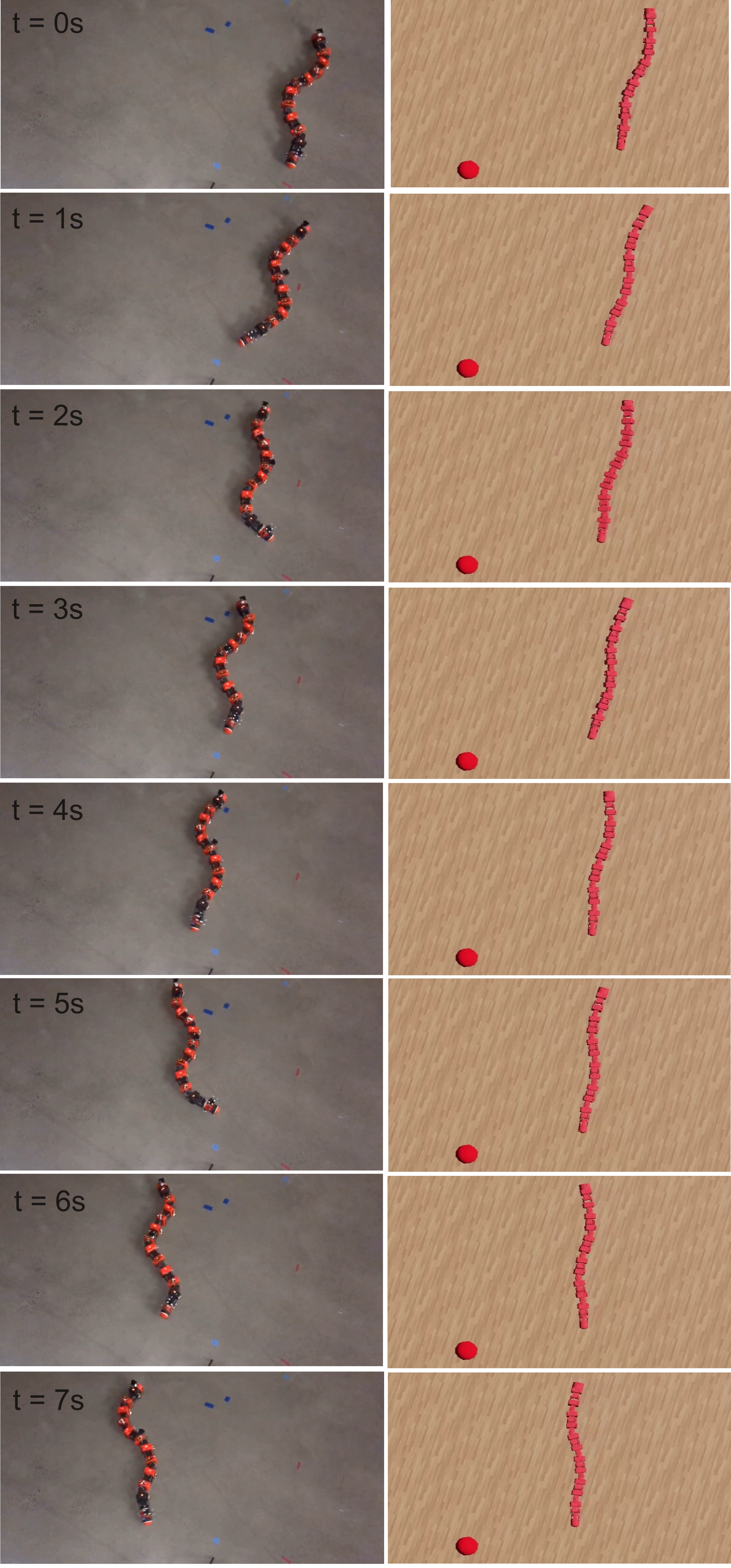} 
  \caption{Highlights the sim2real gap oberserved between COBRA and Webots simulator. The robot performs sidewinding motion in webots with an untuned simulator model. The red ball represents the position attained by the actual robot when executing analogous joint trajectories.}
  \label{fig:disparity}
\end{figure}

The top image in Fig. \ref{fig:disparity} shows the COBRA platform sidewinding on a flat ground. The sidewinding trajectory is generated by the sine equation Eq. \ref{eq:cpg}

\begin{equation}
\begin{aligned}
    y(t) = A \sin(\omega t + \phi)
\end{aligned}
\label{eq:cpg}
\end{equation}
\noindent where \( y(t) \) represents the lateral position of the COBRA platform at time \( t \), \( A \) is the amplitude of the sidewinding motion, \( \omega \) is the angular frequency, and \( \phi \) is the phase angle.

Figure \ref{fig:disparity} hightlights the simulaton-to-reality gap observed in the Webots simulator. Experiments were conducted on the physical robot, and data were collected using motion capture technology to precisely record the trajectories of the head, tail, and middle links. Additionally, joint trajectory data for all eleven joints were recorded. The odd-numbered joints (1, 3, 5, ..., 11 from the head) executed pitching motions, while the even-numbered joints (2, 4, ..., 10 from the head) executed yawing motions. A sidewinding gait was generated by applying a sinusoidal input signal to each joint, serving as a standard input with adjustable parameters to modify the robot's behavior. The input signal followed a sine-wave pattern based on Equation 2, with varying phase differences applied to each joint.
$$
A_\text{pitching} = 14^\circ,~A_\text{yawing} = 60^\circ
$$
$$
\phi = \frac{\pi}{2}[0, 0, 1, 1, 2, 2, 3, 3, 0, 0, 1]
$$

The identical trajectory is also executed within the Webots simulator, allowing us to observe the discrepancy or gap, as shown in Fig. \ref{fig:disparity}, between the simulator and the actual robot. This simulation-to-reality gap hinders the transfer of any learning-based policies from the simulator to the real robot. It is crucial to tackle this issue. The primary objectives of this Master's Thesis is to:

% My strategy to address the sim2real gap involves employing the Reinforcement Learning-based Model Matching approach. This method aims to identify the accurate parameters of COBRA's dynamic model within the simulation environment, thus bridging the gap between simulation and reality. By leveraging these precise parameters, we can develop and train advanced locomotion and manipulation strategies using learning-based techniques. These strategies can then be effectively transferred and implemented on the real robot, enhancing its performance and capabilities.

\begin{itemize}
    \item Develop a framework that reduces the simulation-to-reality gap in the webots simulator for COBRA. This framework will facilitate smoother transitions of policies from the simulator to the actual robot.
    \item Design a controller with the ability to utilize COBRA's locomotion capabilities for manipulating objects within the environment.
\end{itemize}

\begin{figure}
  \centering
  \includegraphics[width=0.9\linewidth]{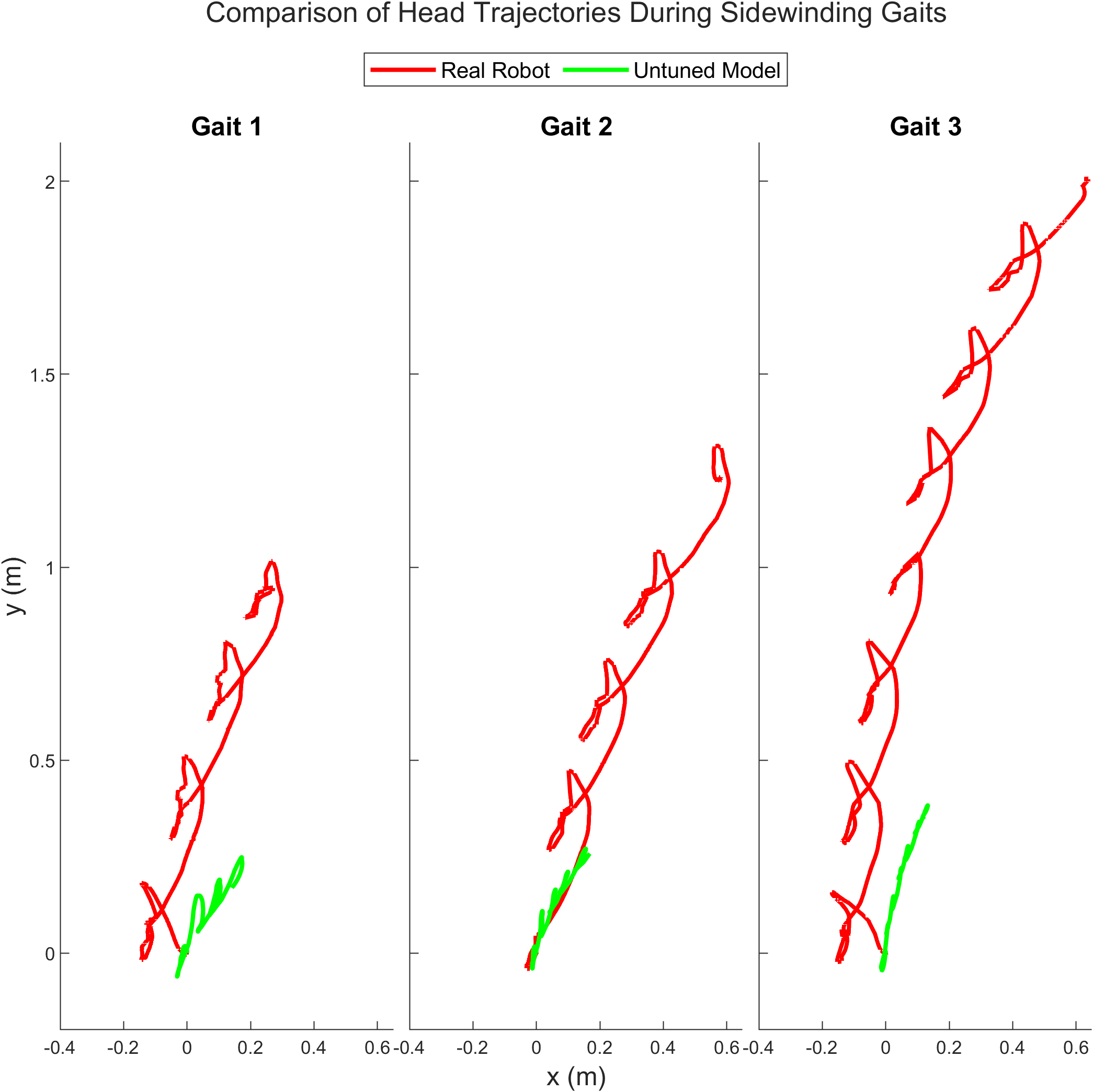} 
  \caption{Illustrates sim-to-real gap observed in COBRA's head link trajectory while performing sidewinding gaits at frequencies of 0.35 Hz, 0.5 Hz, and 0.65 Hz respectively}
  \label{fig:sim2real_head_trajectories}
\end{figure}

\begin{figure}
  \centering
  \includegraphics[width=0.9\linewidth]{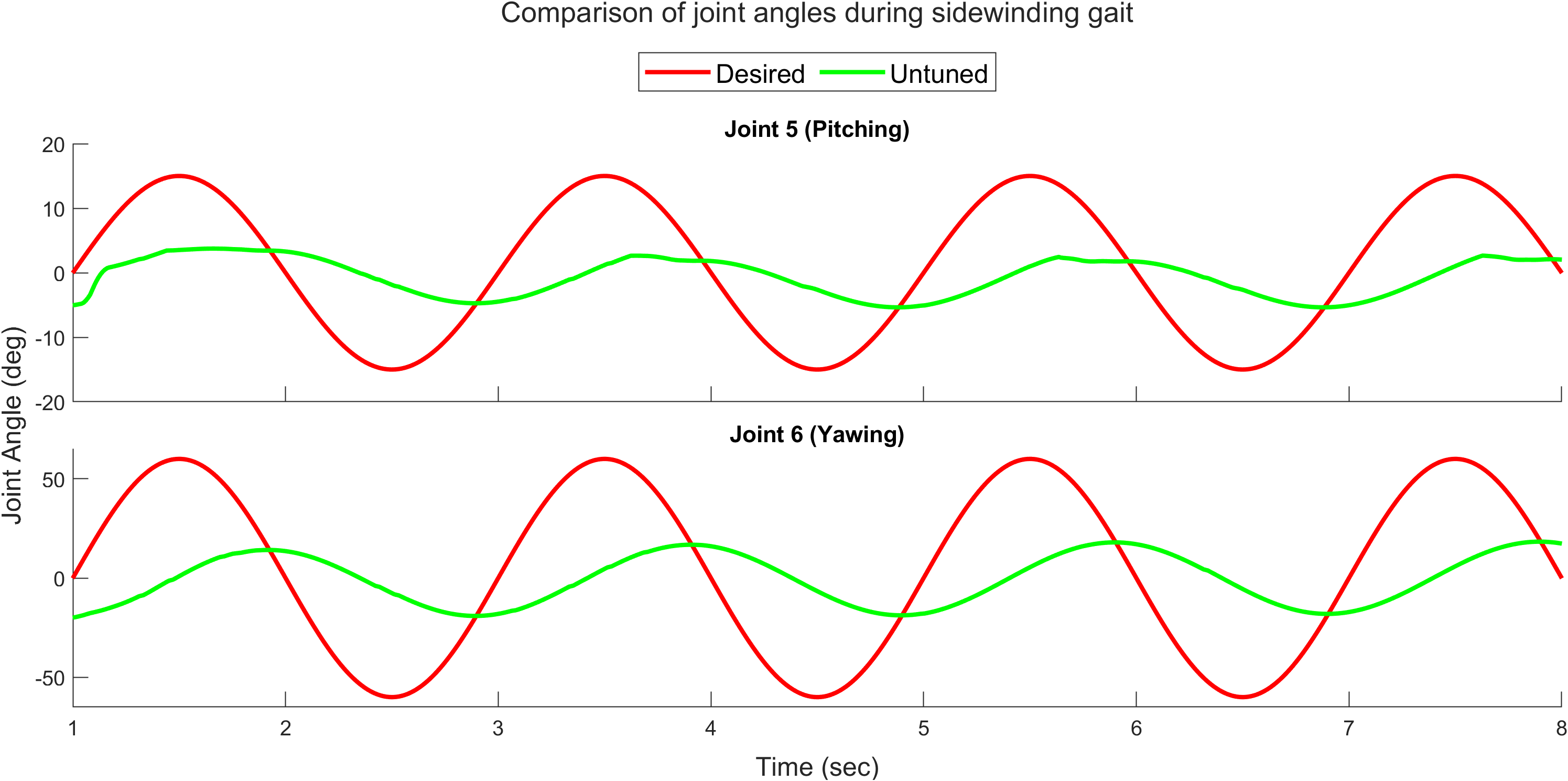} 
  \caption{Illustrates sim-to-real gap observed in COBRA's joint angles while performing sidewinding gaits at frequencies of 0.35 Hz, 0.5 Hz, and 0.65 Hz respectively}
  \label{fig:sim2real_joint_angles}
\end{figure}

This disparity is more evident in Fig. \ref{fig:sim2real_head_trajectories}, where I compare the head trajectories over time for three different gait patterns in both the actual robot and the base simulator (untuned model). Gait 1 represents a sidewinding motion at 0.35 Hz, Gait 2 is sidewinding at 0.5 Hz, and Gait 3 is a sidewinding motion at 0.65 Hz. Additionally, Fig. \ref{fig:sim2real_joint_angles} offers insights into how the performance of actuator model also contributes to the gap observed between simulation and reality. These disparities hinder the seamless transfer of control policies developed in simulation to the real robot, limiting the effectiveness of learned behaviors in practical applications. Hence, addressing the sim-to-real issue is crucial for ensuring the reliability and robustness of COBRA when deployed in diverse and dynamic environments.

% The next major objective of the thesis is to train locomotion policies within a realistically tuned simulator environment for goal-following locomotion tasks. By leveraging advanced simulation techniques and RL algorithms, the thesis aims to develop and optimize locomotion policies that exhibit adaptability, robustness, and efficiency in navigating across a flat surface. The emphasis is on creating locomotion policies that are not only effective in simulation but also transferable and performant when deployed on the actual COBRA robot, thereby bridging the Sim-to-Real gap effectively. Furthermore, the thesis endeavors to introduce a novel framework for loco-manipulation using a modified pure pursuit algorithm. This segment showcases the integration of trained locomotion policies with dynamic manipulation tasks, highlighting COBRA's capabilities in performing complex locomotion and manipulation tasks seamlessly. Through the discussion of challenges and limitations encountered during the training of locomotion policies in accurate simulator parameters, the thesis aims to pave the way for ongoing research efforts and future directions aimed at enhancing COBRA's overall performance, adaptability, and versatility in diverse environments and scenarios.

Chapter \ref{chap:litreview} provides a background on snake robot locomotion principles, Central Pattern Generators (CPGs), and RL agents, setting the foundation for understanding the methodologies used in subsequent chapters. Chapter \ref{chap:sim} delves deeply into the simulation solvers utilized within the simulator, providing a comprehensive analysis of their advantages and disadvantages. Chapter \ref{chap:modelmatching} introduces the COBRA (Crater Observing Bio-inspired Rolling Articulator) snake robot platform, its history, and motivation and delves into the model matching framework developed to bridge the simulation-to-reality gap. This framework utilizes RL agents to accurately tune simulation parameters for effective sim-to-real transfer. Chapter \ref{chap:locomanipulation} introduces the loco-manipulation controller framework using a modified pure pursuit algorithm and discusses its integration with trained locomotion policies for dynamic manipulation tasks. Experimental validation and results are presented in Chapter \ref{chap:results}, showcasing the results of the developed methodologies. Chapter \ref{chap:conclusion}, talks about conclusions of the work along with discussions on future research directions.

\section{Contributions}

My thesis contributes in laying the foundation for solving the challenging sim2real problem on the COBRA platform and developing advanced locomotion strategies such as loco-manipualtion. The following is a detailed list of my contributions:

\begin{itemize}
    \item Created an API for the Webots simulator, allowing access to the simulator model parameters for a downstream pipeline.
    \item Developed a model matching framework using an RL agent to accurately determine simulator model parameters, thereby minimizing the simulation-to-reality gap between webots simulator and the real robot.
    \item Formulated a modified pure pusuit algorithm for the purpose of real-time directed locomotion on COBRA.
    \item Introduced a framework for loco-manipulation using the modified pure pursuit algorithm, showcasing COBRA's capabilities in performing complex loco-manipulation task seamlessly.
    \item Assisted in realizing open-loop gait patterns for the purpose of loco-manipulation on the COBRA platform.
\end{itemize}

%% litreview
\chapter{Literature Review}
\label{chap:litreview}

%\iffalse
%\begin{figure}[h!]
 %   \centering
 %   \includegraphics[\linewidth=1.0]{fig/husy_morph.png}
 %   \caption{Illustrates Northeastern University’s Husky Carbon Platform, which is a multi-modal %legged-aerial robot \textcolor{red}{ToDo: Change to high quality render of Husky }}
  %  \label{fig:husky}
%\end{figure}
%\fi

This chapter provides a comprehensive overview of advancements in robotic control and learning strategies, focusing on \ac{CPG}, Learning-based Strategies for Locomotion Tasks, and Simulation-to-Reality Transfer of Trained Policies. It highlights the versatility and robustness of \ac{CPG} in generating rhythmic behaviors for snake-like and legged robots, as well as their integration with deep learning for adaptive control and efficient learning. Additionally, the chapter discusses reinforcement learning as a powerful paradigm for developing advanced locomotion strategies in various robotic platforms, showcasing improved navigation, adaptability, and robustness across different terrains and morphologies. Furthermore, it explores innovative approaches to transferring trained policies from simulation to real-world environments, emphasizing the evolving landscape of simulation-to-reality transfer in robotics and machine learning.

\section{Central Pattern Generators}

\ac{CPG} are neural-inspired controllers widely used for generating rhythmic behaviors in robotic systems, particularly in those requiring oscillatory motions like snake-type or legged robots. The application of \ac{CPG} is driven by their versatility and robustness to uncertainties, making them suitable for real-world tasks. \cite{garcia-saura_central_2015} focuses on locomotion control in snake-like robots with cardan joints using a CPG approach. The developed \ac{CPG} model comprises a double chain structure of nonlinear oscillators interconnected with diffusive couplings. This model demonstrates stability in producing rhythmic patterns for serpentine and sidewinding locomotion. The paper also discusses explicit control parameters within the \ac{CPG} model, such as oscillation frequencies, amplitudes, and phase differences between oscillators, allowing for speed and direction adjustments during locomotion. Simulation and experimental results on real snake robots confirm the successful control achieved using the proposed \ac{CPG} approach.

Another research effort \cite{deshpande_deepcpg_2023} introduces DeepCPG policies, embedding CPGs into a neural network layer for end-to-end learning of locomotion behaviors in multi-legged robots. These policies facilitate sample-efficient learning even with high-dimensional sensor spaces, such as vision inputs. The study showcases successful locomotion strategies learned through DeepCPG policies, emphasizing the potential for scaling these policies in modular robot configurations and multiagent setups. Experimental results demonstrate effective sensor-motor integration based on biological principles, highlighting the adaptability and complexity achievable through DeepCPG approaches. A novel locomotion control method \cite{liu_learning_2020} for soft robot snakes combines \ac{DRL} for adaptive goal-tracking behaviors with a CPG system utilizing Matsuoka oscillators. The integrated control architecture includes an RL module regulating inputs to the CPG system, which in turn translates outputs into pressure inputs for pneumatic actuators. This setup allows for independent control of oscillation frequency and amplitude, enhancing learning performance and data efficiency. Experimental validation on simulated and real soft snake robots validates the effectiveness of the proposed adaptive controller. These studies \cite{wang_cpg-inspired_2017} \cite{zhenli_lu_serpentine_2005} \cite{manzoor_serpentine_2020} collectively showcase the diverse applications and advancements in CPG-based models for controlling various t  ypes of robotic systems, emphasizing their adaptability, robustness, and potential for real-world implementations.

\section{Learning-based Strategies for Locomotion Tasks}

Reinforcement Learning \cite{schulman_proximal_2017} \cite{radosavovic_learning_2023} has emerged as a powerful paradigm for developing advanced robotics strategies, offering solutions that can adapt and learn in complex and dynamic environments. This section reviews recent developments in RL-based robotics strategies, encompassing various applications and methodologies.

One study \cite{lee_learning_2020} challenges the traditional approach of local navigation for legged robots, advocating for an end-to-end policy trained with deep reinforcement learning. By focusing on reaching a target position within a specified time rather than continuous path tracking, this approach enables the robot to explore a broader solution space and learn more complex behaviors. The successful deployment of policies on a real quadrupedal robot demonstrates improved navigation across challenging terrains while conserving energy and achieving higher success rates.

Another framework \cite{feng_genloco_2022} introduces Generalized Locomotion (GenLoco) controllers for quadrupedal robots, aiming to synthesize controllers that can be applied across a range of similar robot morphologies. By training on a diverse set of simulated robots, these controllers acquire more general control strategies, leading to successful transfers to novel robots with varied morphologies in both simulated and real-world settings. 

A novel RL framework \cite{xie_feedback_2018} is proposed for designing feedback control policies for 3D bipedal walking, leveraging physical insights and hybrid zero dynamics approaches. This framework demonstrates stable limit walking cycles capable of tracking different walking speeds and resisting adversarial forces, showcasing robust and efficient bipedal locomotion control. A hierarchical framework \cite{ouyang_adaptive_2021} is presented for goal-oriented locomotion, dividing the problem into learning primitive skills and utilizing model-based planning for sequencing these skills. This approach improves sample efficiency and generalizability, enabling robots to navigate to arbitrary goals with sensory noise, demonstrating efficacy on hardware platforms such as Daisy hexapod. 

A fully learning-based method \cite{radosavovic_learning_2023} for real-world humanoid locomotion is introduced, utilizing a causal Transformer model trained through large-scale RL on randomized environments. This approach eliminates the need for state estimation, trajectory optimization, or pre-computed gait libraries, achieving successful deployment on a real robot for full-sized humanoid locomotion. A radically robust controller \cite{escontrela_zero-shot_2020} is developed for legged locomotion in challenging natural environments, trained via RL in simulation and exhibiting zero-shot generalization to diverse terrains. This controller, based on neural networks and proprioceptive feedback, surpasses previous work in robustness and generalizability, showcasing remarkable adaptability in real-world scenarios. 

A training setup using massive parallelism \cite{rudin_learning_2022} on a single workstation GPU is explored for fast policy generation in real-world robotic tasks. The impact of different training algorithm components on policy performance and training times is analyzed, achieving significant speedups compared to previous methods. A framework for learning reward functions \cite{daniel_active_2014} through active learning, querying human expert knowledge for a subset of the agent's rollouts, is proposed. This method demonstrates improved reward learning and generalization on tasks such as grasping, addressing challenges in defining reward functions manually. These RL-based robotics strategies represent advancements in autonomous control, navigation, and adaptability, showcasing the potential of RL techniques in enhancing robotic capabilities across various applications and environments.

\section{Simulation-to-Reality Transfer of Trained Policies}

In the realm of robotics and machine learning, the challenge of transferring trained policies from simulation environments to the real world has been a persistent hurdle. This literature review synthesizes and analyzes several papers that address this problem from different angles, shedding light on innovative approaches and advancements in the field.

One significant approach discussed in the literature is the adaptation of simulation parameter distributions based on real-world roll-outs during policy training. The paper by \cite{chebotar_closing_2019} introduces this concept, emphasizing the importance of matching policy behavior between simulation and reality through dynamic adjustment of simulation parameters. By leveraging a distribution of simulated scenarios and iteratively adapting parameters using real-world data, policies trained with this method exhibit reliable transferability to different robots across tasks such as swing-peg-in-hole and opening a cabinet drawer. Another critical aspect highlighted in the literature is domain randomization, a technique aimed at bridging the gap between simulation and real-world environments. \cite{mehta_active_2019} underscores the potential of \ac{DRL} algorithms in controlling real robots by training policies on a diverse distribution of environmental conditions in simulation. This diversity enables policies to generalize effectively to real-world scenarios, although the optimal design of this distribution remains an open question. Furthermore, the literature explores methods for automatically adjusting simulation parameters to maximize model accuracy.

\cite{vuong_how_2019} presents a reinforcement learning-based approach that dynamically controls simulator parameters to optimize data synthesis for model training. Unlike previous techniques that relied on hand-crafted parameter settings, this method autonomously converges to optimal simulation parameters, enhancing the accuracy of trained models. A notable contribution in the domain of locomotion controllers for bipedal robots is the work by \cite{yu_sim--real_2019}, which introduces a model-free reinforcement learning framework combined with domain randomization for sim-to-real transfer. By encouraging policies to learn robust behaviors across varied system dynamics, this approach achieves versatile walking behaviors on real bipedal robots, surpassing traditional controllers' robustness. 

\cite{qin_sim--real_2019} adopts an end-to-end approach using an Actor-Critic network with \ac{PPO} to train six-legged robots in simulation and successfully transfer learned models to real hardware. Curriculum learning is employed to accelerate and optimize training, showcasing the potential of \ac{DRL} in enhancing locomotion for legged robots. Moreover, the concept of \ac{SDR} is introduced by , focusing on context-aware placement of objects and distractors in simulated scenes. This approach enhances neural network performance for tasks like 2D bounding box car detection, outperforming other synthetic data generation methods and even real-world data in certain scenarios. The reviewed literature underscores the evolving landscape of simulation-to-reality transfer for trained policies in robotics and machine learning. From adaptive simulation parameter distributions to domain randomization and structured data generation, these approaches collectively contribute to overcoming challenges and advancing the capabilities of models in real-world applications.

\section{Research Analysis}

\begin{figure}
  \centering
  \includegraphics[width=1.0\linewidth]{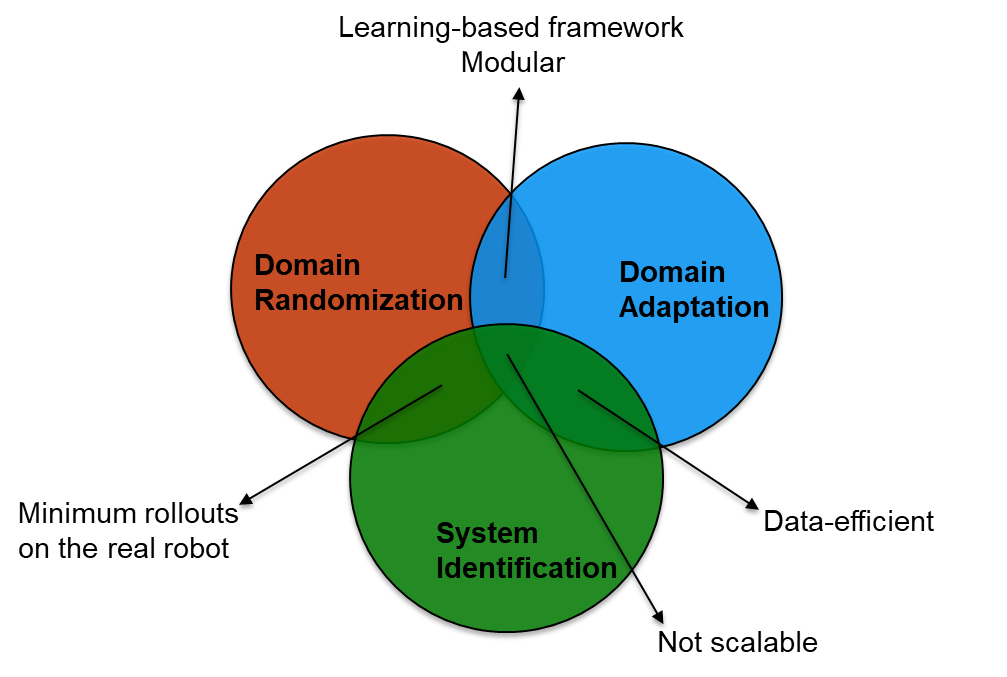} 
  \caption{Highlights the shared aspects among the popular solutions to sim2real problem.}
  \label{fig:research_gap}
\end{figure}

While significant progress has been made in the domain of simulation-to-reality transfer for trained policies in locomotion tasks, several research gaps and challenges remain unaddressed. This section outlines key areas where further investigation and development are needed to advance the field.

One prominent research gap pertains to the need for a framework that bridges the simulation-to-reality gap specifically for locomotion tasks that do not require heavy computation or vast amounts of real-world rollouts to tune simulation model parameters. Existing approaches such as domain randomization and simulation parameter adaptation often rely on computationally intensive processes or extensive data collection on real robots. This limitation hinders their scalability and applicability to scenarios where computational resources or real-world data are limited. Furthermore, there is a gap in developing efficient and scalable methods for sim-to-real transfer in locomotion tasks across diverse robot morphologies and environments. Many current approaches focus on specific robot types or controlled environments, lacking generalizability to varied real-world conditions. A comprehensive framework that can adapt to different robot platforms, terrains, and uncertainties while maintaining transferability and performance is crucial for broader deployment and adoption in real-world applications.

Another research gap lies in the integration of domain knowledge and physical insights into sim-to-real transfer frameworks. While data-driven approaches like reinforcement learning have shown promise, incorporating domain expertise and leveraging physical principles can enhance the robustness and efficiency of transfer learning algorithms. Developing hybrid approaches that combine data-driven techniques with domain knowledge can lead to more effective sim-to-real transfer strategies for locomotion tasks. Additionally, there is a need for benchmarking and standardization in evaluating sim-to-real transfer methods for locomotion. Currently, the assessment of transferability and generalization often lacks standardized metrics and evaluation protocols, making it challenging to compare different approaches objectively. Establishing benchmark datasets, evaluation criteria, and performance metrics tailored specifically for locomotion tasks can facilitate fair comparisons and drive innovation in sim-to-real transfer research.

The research gaps identified emphasize the necessity for a framework that addresses the sim-to-real gap in locomotion tasks efficiently, scalably, and with consideration for diverse robot morphologies, environments, and domain knowledge integration. By addressing these gaps, researchers can pave the way for more robust, adaptable, and widely applicable sim-to-real transfer solutions in the field of robotics and machine learning.

\chapter{Simulators and Physics Engines}
\label{chap:sim}

This chapter delves into the critical role of simulation environments in the development and evaluation of \ac{DRL}-based gait design for robots. Simulators play a pivotal role by providing a virtual arena where researchers and engineers can explore different scenarios, train control policies, and assess robot behavior without the need for physical hardware. This chapter explores several prominent simulators used in robotics research, including Webots, Simscape, Gazebo, PyBullet, MuJoCo, and Isaac Sim. Each simulator utilizes distinct simulation solvers, such as the \ac{ODE} in Webots, Simscape's ODE4 equation and stick-slip friction model, Gazebo's integration with \ac{ODE}, PyBullet's Bullet Physics SDK, MuJoCo's custom physics engine based on \ac{ODE}, and Isaac Sim's LULA solver leveraging NVIDIA's \ac{GPU}-accelerated physics simulation. Understanding these solvers' features, capabilities, and limitations is crucial for selecting the most suitable simulator for specific research goals and ensuring accurate and realistic RL-based locomotion strategy development.

\section{Physics Engines}
\subsection{Open Dynamics Engine}

\begin{table}[htbp]
    \centering
    \caption{Comparison of simulators}
    \label{tab:sim_talbe}
    \begin{tabular}{|p{2cm}|p{4cm}|p{4cm}|p{4cm}|}
        \hline
        \textbf{Simulator} & \textbf{Physics Engine} & \textbf{Pros} & \textbf{Cons} \\
        \hline
        Simscape & Multibody dynamics & High-fidelity simulations & Incompatible with RL agent training \\
        \hline
        Webots & Open dynamics engine & Compatible with RL agent training & Accuracy of simulations \\
        \hline
        PyBullet & Bullet physics SDK & Ease of integration with deep learning frameworks & Limitations in friction modeling \\
        \hline
        MuJoCo & Open dynamics engine & Popular for training locomotion-based policies & Accuracy of simulations \\
        \hline
        Gazebo & Open dynamics engine & Tight integration with ROS & Limitations in modeling contact dynamics \\
        \hline
        Isaac Sim & PhysX & Massive parallelization in RL training & Requires access to compatible hardware \\
        \hline
    \end{tabular}
\end{table}

\ac{ODE} excels in simulating articulated rigid body structures, where rigid bodies are connected by various joints, such as those found in ground vehicles (like wheels to the chassis), legged creatures (like legs to the body), or stacks of objects. It is specifically tailored for interactive or real-time simulations, making it ideal for dynamic virtual reality environments with changing conditions. Its strengths lie in being fast, robust, and stable, allowing users the flexibility to modify the system's structure during simulation.

One of ODE's key features is its use of a highly stable integrator, preventing simulation errors from spiraling out of control and avoiding sudden system "explosions" that can occur in other simulators. ODE prioritizes speed and stability over absolute physical accuracy, making it well-suited for applications where real-time performance is crucial.

ODE implements hard contacts, employing non-penetration constraints when bodies collide, unlike other simulators that may rely on virtual springs for contact representation, a method prone to errors. Additionally, ODE includes a built-in collision detection system with primitives such as spheres, boxes, cylinders, capsules, planes, rays, and triangular meshes, allowing for fast identification of potential intersections using collision "spaces."

Key features of ODE include support for:
\begin{itemize}
    \item Rigid bodies with customizable mass distribution.
    \item Various joint types like ball-and-socket, hinge, slider (prismatic), hinge-2, fixed, angular motor, linear motor, and universal joints.
    \item Collision primitives including spheres, boxes, cylinders, capsules, planes, rays, triangular meshes, and convex shapes.
    \item Different collision spaces like Quad trees, hash spaces, and simple spaces for efficient collision detection.
    \item Simulation methods based on Lagrange multiplier velocity models by Trinkle/Stewart and Anitescu/Potra, using a first-order integrator (higher-order integrators planned for future updates).
    \item Choice of time stepping methods such as the standard "big matrix" method or the newer iterative QuickStep method.
    \item Contact and friction models based on the Dantzig LCP solver, providing a faster approximation of the Coulomb friction model.
    \item Native C interface with additional C++ interfaces for ease of use and integration.
    \item Extensive unit testing and platform-specific optimizations for enhanced performance.
\end{itemize}

Additionally, ODE incorporates constraint parameters that play a crucial role in simulating realistic interactions between objects. Parameters like \ac{CFM} and \ac{ERP} influence how constraints are enforced and how joint errors are handled during simulation steps. These parameters allow users to control the stability and accuracy of simulations, especially in scenarios with intermittent contacts or complex interactions. Overall, understanding and manipulating these inner workings of the ODE engine enable users to create more realistic and effective simulations for various applications in robotics and virtual environments.

\subsubsection{Constraint Force Mixing}

A joint group serves as a specialized container within the ODE framework that holds joints within a world. Joints can be assigned to a group, and when the joints are no longer needed, the entire group can be swiftly destroyed with a single function call. However, individual joints within a group cannot be removed until the entire group is emptied. This feature is particularly advantageous in scenarios involving contact joints, where groups of joints are added and removed from the world during each time step.

In the context of joint error and the ERP, when a joint connects two bodies, these bodies must maintain specific positions and orientations relative to each other. Joint error can occur due to two main reasons: improper setting of one body's position/orientation without corresponding adjustments to the other body, or errors introduced during simulation leading to deviations from the required positions. To mitigate joint error, ODE applies a special force during each simulation step to realign the bodies, controlled by the ERP, which ranges from 0 to 1. ERP determines the proportion of joint error corrected in the next time step, with values closer to 1 attempting to fix all error but not recommended due to practical approximations, while values between 0.1 and 0.8 are commonly used.

The distinction between hard and soft constraints is crucial in ODE's constraint handling. While most constraints are inherently "hard," meaning they must never be violated (e.g., a ball always remains in a socket), soft constraints allow controlled violations, simulating softer materials. \ac{CFM} parameter plays a key role in this distinction. CFM mixes the resulting constraint force with the original constraint equation, allowing controlled violations proportional to CFM times the restoring force needed to enforce the constraint. Positive CFM values enable soft constraints, while negative values are discouraged due to potential instability.

ERP and CFM can be independently adjusted in various joints to control their stiffness and compliance. For instance, higher CFM values introduce more "softness" to the constraint, while adjusting ERP and CFM allows simulating spring-like or spongy behaviors. Increasing global CFM can enhance simulation stability and reduce numerical errors, especially in near-singular systems. Additionally, collision handling in ODE involves pre-simulation collision detection to identify contact points, creation of specialized contact joints, grouping these joints for efficient management, and removal of contact joints post-simulation.

Mathematically, the relationship between ERP, CFM, and the desired spring and damper constants can be expressed as:

\begin{equation}
\begin{aligned}
    \text{ERP} = \frac{h \cdot k_p}{h \cdot k_p + k_d} \\
    \text{CFM} = \frac{1}{h \cdot k_p + k_d} \\
\end{aligned}
\label{eq:erf_cpm}
\end{equation}

where \( h \) represents the step size, \( k_p \) is the spring constant, and \( k_d \) is the damping constant, ensuring equivalence between ODE's parameters and those of a spring-and-damper system simulated using implicit first-order integration.

Traditionally, the constraint equation for every joint has the form

\begin{equation}
\begin{aligned}
    Jv=c \\
\end{aligned}
\label{eq:coneq}
\end{equation}

where \( v \) is a velocity vector for the bodies involved, \( J \) is a "Jacobian" matrix with one row for every degree of freedom the joint removes from the system, and \( c \) is a right-hand side vector. At the next time step, a vector \( \lambda \) is calculated (of the same size as \( c \)) such that the forces applied to the bodies to preserve the joint constraint are:

\begin{equation}
\begin{aligned}
    Fc=J^T\lambda \\
\end{aligned}
\label{eq:Fc}
\end{equation}

ODE adds a new twist. ODE's constraint equation has the form

\begin{equation}
\begin{aligned}
    Jv=c+CFM\lambda \\
\end{aligned}
\label{eq:ode_const}
\end{equation}

where CFM is a square diagonal matrix. CFM mixes the resulting constraint force in with the constraint that produces it. A nonzero (positive) value of CFM allows the original constraint equation to be violated by an amount proportional to CFM times the restoring force \( \lambda \) that is needed to enforce the constraint. Solving for \( \lambda \) gives

\begin{equation}
\begin{aligned}
     (J M^{-1} J^T + \frac{1}{h} CFM) \lambda = \frac{1}{h} c \\
\end{aligned}
\label{eq:CFM}
\end{equation}

Thus, CFM simply adds to the diagonal of the original system matrix. Using a positive value of CFM has the additional benefit of taking the system away from any singularity and thus improving the factorizer accuracy.

\subsubsection{Friction Approximation}

\begin{figure}
  \centering
  \includegraphics[width=0.5\linewidth]{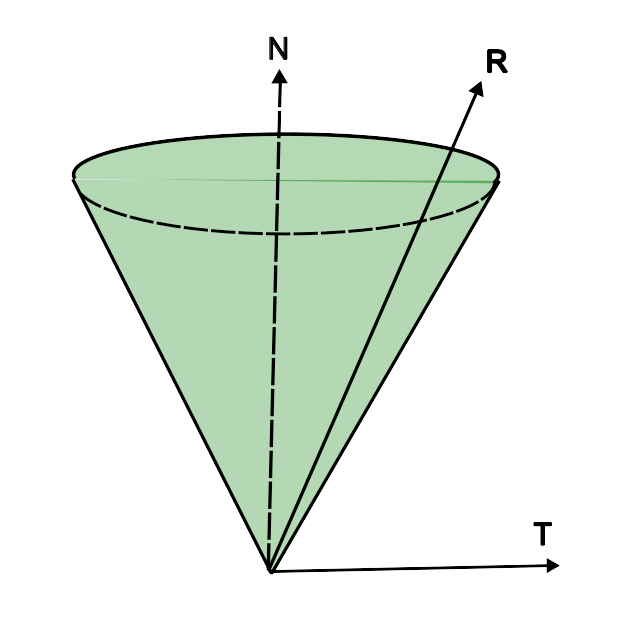} 
  \caption{Illustrates friction cone inside the simulator}
  \label{fig:friction_cone}
\end{figure}

The Coulomb friction model offers a straightforward yet efficient approach to depicting friction at contact points. It establishes a relationship between normal and tangential forces at a contact point, denoted as FT and FN respectively, with $\mu$ representing the friction coefficient (usually around 1.0). The model is defined by the inequality $|FT| \leq \mu|FN|$, creating a "friction cone" visualization where FN serves as the cone's axis and the contact point as its vertex. When the total friction force vector aligns within this cone, the contact remains in "sticking mode," preventing relative movement between surfaces. Conversely, if the force vector lies on the cone's surface, the contact transitions to "sliding mode," allowing sliding between surfaces. The parameter $\mu$ dictates the maximum ratio of tangential to normal force.

ODE's friction models are approximations of this friction cone, primarily for efficiency purposes. Two main approximations are available:
\begin{itemize}
    \item Altering $\mu$ to denote the maximum tangential force at a contact in either tangential direction, independent of the normal force. Although somewhat non-physical as it disregards normal force, it is computationally economical and allows for a force limit suitable for simulation purposes.
    \item Approximating the friction cone with a friction pyramid aligned along the first and second friction directions. This method first computes normal forces assuming frictionless contacts, calculates maximum limits for tangential forces ($Fm = \mu|FN|$), and then solves the system using these limits. While not a true friction pyramid, this approximation is easier to implement as $\mu$ remains unit-less and can be set as a constant around 1.0 without specific simulation considerations.
\end{itemize}

\subsection{Simscape Multibody Dynamics}

Simscape facilitates the swift creation of models for physical systems within the Simulink environment. Through Simscape, users construct physical component models utilizing direct physical connections that seamlessly integrate with block diagrams and other modeling approaches. These models encompass various systems like electric motors, bridge rectifiers, hydraulic actuators, and refrigeration systems, constructed by assembling fundamental components into a schematic. Additional Simscape add-on products offer more intricate components and analysis functionalities.

\begin{figure}
  \centering
  \includegraphics[width=0.5\linewidth]{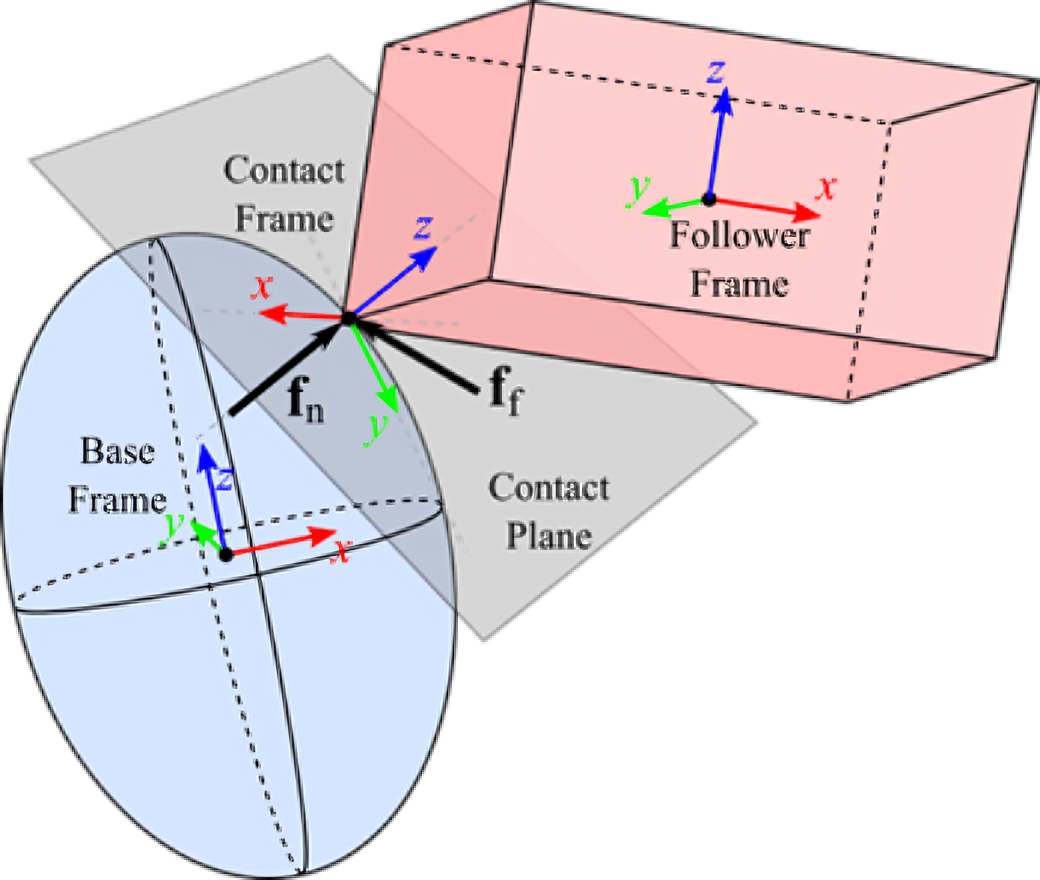} 
  \caption{Illustrates dynamic forces in multibody models, such as normal and frictional forces. These forces influence the behavior of solid blocks in contact \cite{noauthor_simscape_nodate}.}
  \label{fig:simscape_mutibody}
\end{figure}

Simscape aids in the development of control systems and the assessment of system-level performance. Users can craft custom component models using the MATLAB-based Simscape language, allowing for text-based creation of physical modeling components, domains, and libraries. Models can be parameterized using MATLAB variables and expressions, and control systems can be designed within Simulink for the physical system. Simscape also supports C-code generation to deploy models to other simulation environments, including hardware-in-the-loop (HIL) systems.

When simulating interactions between solid blocks in contact, the influence of contact forces becomes crucial in determining their behaviors. Both the normal force (\( f_n \)) and the frictional force (\( f_f \)) can significantly impact the dynamic responses of a multibody model. Contact forces are particularly relevant in various modeling scenarios, including package conveyors, robotic movement, and race car dynamics. The Spatial Contact Force block in multibody dynamics simulation represents forces between solid bodies' base and follower frames. This block applies equal and opposite forces along a shared contact plane when two solid blocks are linked, adhering to Newton's Third Law. The normal force is calculated based on penetration depth and velocity, while the frictional force, if applied, depends on the normal force and relative velocities at the contact point.

The Spatial Contact Force block features three expandable nodes for setting its properties: Normal Force, Frictional Force, and Sensing. We utilize these properties when employing the Spatial Contact Force block to model contact forces between two solid bodies.

\textbf{Normal Force: }The parameters within the Normal Force section determine the normal force (\( f_n \)) exerted by the two solid bodies on each other. You can define the Stiffness, Damping, and Transition Region Width.

\textbf{Frictional Force: }The parameters within the Frictional Force section determine the frictional force (\( f_f \)) exerted by the two solid bodies on each other. If you select the Smooth Stick-Slip Method, you can specify the Coefficient of Static Friction, Coefficient of Dynamic Friction, and Critical Velocity. Note that these options are not available when the Method is set to None.

\begin{figure}
  \centering
  \includegraphics[width=0.8\linewidth]{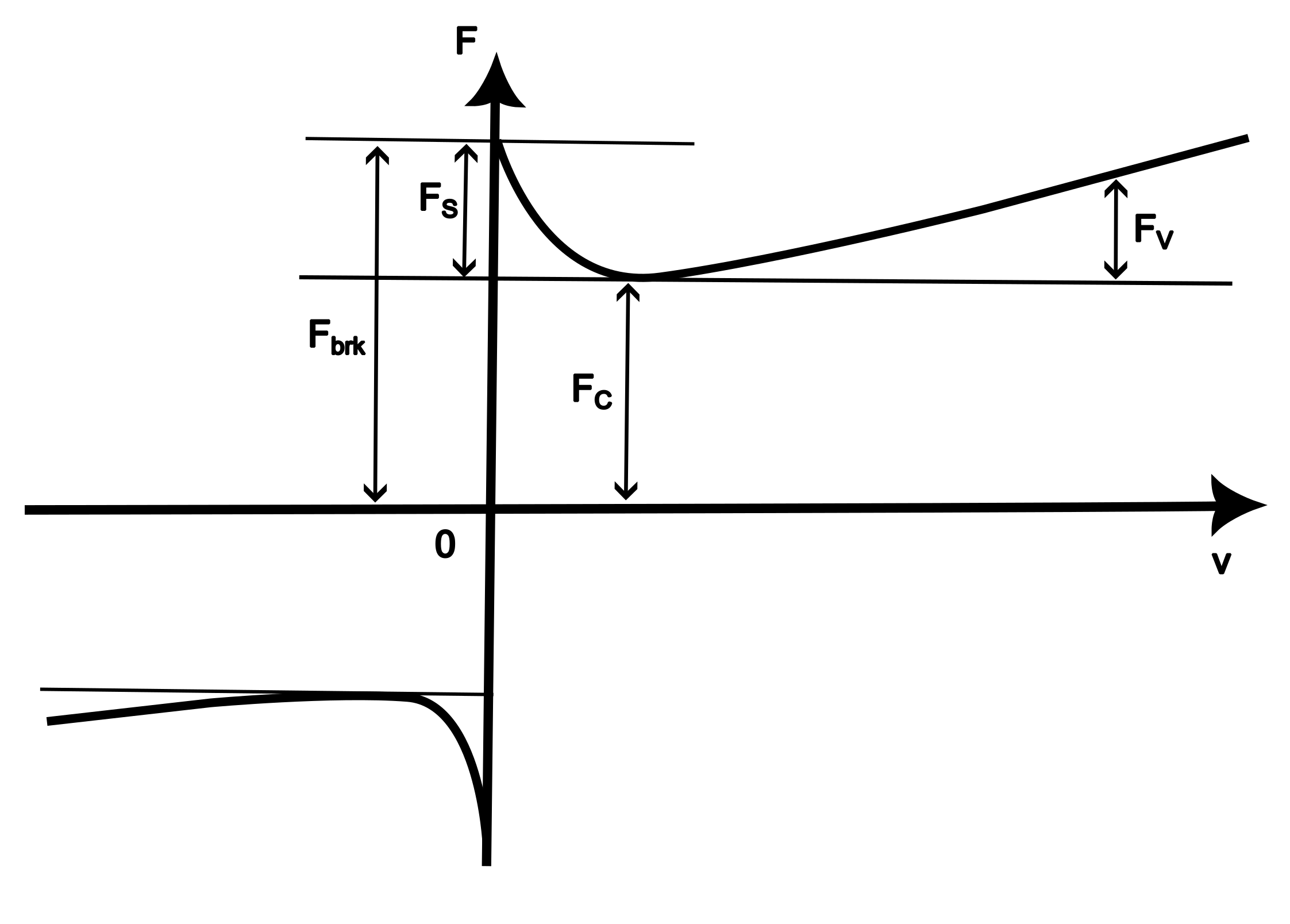} 
  \caption{Illustrates the friction model used inside simscape simulator \cite{noauthor_httpswwwmathworkscomproductssimscapehtml_nodate}.} 
  \label{fig:friction_model}
\end{figure}

The Translational Friction block represents friction occurring between moving bodies, with the friction force being a function of their relative velocity. It includes Stribeck, Coulomb, and viscous components, as depicted in the accompanying figure. Stribeck friction (FS) exhibits a negatively sloped characteristic at low velocities, while Coulomb friction (FC) remains constant regardless of velocity. Viscous friction (FV) opposes motion proportionally to the relative velocity. Near zero velocity, the sum of Coulomb and Stribeck frictions is termed breakaway friction (Fbrk). The friction can be approximated using the following equations:

\begin{equation}
\begin{aligned}
     F = \sqrt{2e(F_{brk} - F_{C}) \cdot \exp\left(-\left(\frac{v}{v_{St}}\right)^2\right) \cdot \frac{v}{v_{St}} + F_{C} \cdot \tanh\left(\frac{v}{v_{Coul}}\right) + f_v} \\
\end{aligned}
\label{eq:friction_1}
\end{equation}

\begin{equation}
\begin{aligned}
    v_{St} = \frac{v_{brk}}{\sqrt{2}} \\
    v_{Coul} = \frac{v_{brk}}{10} \\
    v = v_R - v_C \\
\end{aligned}
\label{eq:friction_2}
\end{equation}

where \( F \) is the friction force, \( F_C \) is Coulomb friction, \( F_{brk} \) is breakaway friction, \( v_{brk} \) is breakaway friction velocity, \( v_{St} \) is Stribeck velocity threshold, \( v_{Coul} \) is Coulomb velocity threshold, \( v \) is relative velocity, \( f_v \) is the viscous friction coefficient.

The exponential function in the Stribeck portion ensures continuity and decays for velocities greater than breakaway friction velocity. The hyperbolic tangent function in the Coulomb portion maintains smoothness and continuity at zero velocity but quickly reaches its full value at nonzero velocities. The block's positive direction, from port R to port C, implies force transmission from R to C when port R's velocity exceeds that of port C.

\section{Webots Simulator for COBRA}

I chose the Webots simulator as the ideal platform for bridging the sim2real gap in training reinforcement learning (RL) agents for the COBRA platform due to its seamless integration capabilities with RL agent training and its extensive resources for developing locomotion, manipulation, and navigation controllers/policies. Webots offers a user-friendly interface and supports various programming languages like Python and C++, making it easy to implement and integrate RL algorithms. Its diverse library of robot models, realistic physics simulation, and visualization tools provide a conducive environment for training RL-based locomotion policies. Additionally, Webots provides resources and tools for developing controllers and policies for various tasks, including locomotion, manipulation, and navigation, which are essential for the COBRA platform's functionalities. Overall, Webots' compatibility with RL training and its comprehensive resources make it an excellent choice for addressing the sim2real gap in developing efficient controllers for the COBRA platform.

\chapter{Model Matching Framework}
\label{chap:modelmatching}

\section{COBRA Platform}
\begin{figure}
  \centering
  \includegraphics[width=1.0\linewidth]{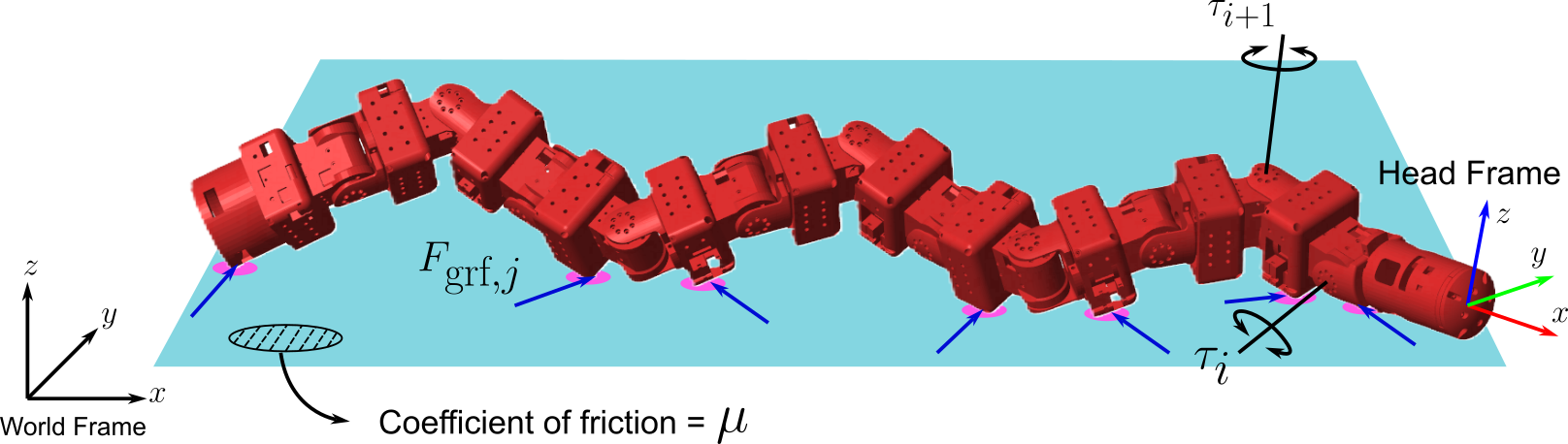} 
  \caption{Illustrates free body diagram of the \ac{COBRA} platform}
  \label{fig:fbd}
\end{figure}

COBRA, an acronym for Crater Observing Bio-inspired Rolling Articulator, is a snake robot designed after the intricate locomotion of serpentine creatures \cite{salagame_how_2023} \cite{rollinson_gait-based_2013} \cite{mori_reinforcement_nodate}. This robot is specifically engineered to navigate and endure rugged terrains and environments, such as craters, where traditional wheeled or legged robots may struggle. Figure \ref{fig:fbd} shows a free body diagram of the COBRA platform and the ground coefficients of friction. With 11 joints and 12 links, the COBRA platform exhibits a complex mechanical structure that allows for versatile movements and functionalities. The presence of 6 yawing joints and 5 pitching joints enables the platform to achieve a wide range of orientations and configurations, making it suitable for diverse applications in robotics and engineering.

The free body diagram provides a detailed representation of the forces and moments acting on the COBRA platform as it interacts with the ground. The coefficients of friction play a crucial role in analyzing the stability and motion dynamics of the platform during various maneuvers.

\begin{equation}
    \begin{aligned}
        \left[\begin{array}{cc}
D_H & D_{H a} \\
D_{a H} & D_a
\end{array}\right]\left[\begin{array}{l}
\ddot{q}_H \\
\ddot{q}_a
\end{array}\right]&+\left[\begin{array}{l}
H_H \\
H_a
\end{array}\right]=\left[\begin{array}{l}
0 \\
B_a
\end{array}\right]u+\\&
+
\left[\begin{array}{cc}
J_{H}^\top \\ J_{a}^\top
\end{array}\right]F_{GRF}
    \end{aligned}
\label{eq:full-dynamics}
\end{equation}
\noindent where $D_i$, $H_i$, $B_i$, and $J_i$ are partitioned model parameters corresponding to the head 'H' and actuated 'a' joints. $F_{GRF}$ denotes the ground reaction forces. $u$ embodies the joint actuation torques. The actuator parameters include: transmission inertia, internal damping, and dc motor constant. The reference trajectories used in the actuator models are generated by 
% \begin{equation}
% \begin{aligned}
%     y(t) = A \sin(\omega t + \phi)
% \end{aligned}
% \label{eq:cpg}
% \end{equation}
% \noindent where \( y(t) \) represents the lateral position of the COBRA platform at time \( t \), \( A \) is the amplitude of the sidewinding motion, \( \omega \) is the angular frequency, and \( \phi \) is the phase angle.

The sidewinding motion depicted in Fig. \ref{fig:disparity} demonstrates the platform's capability to traverse terrain with varying obstacles and inclines. By utilizing a sine-based trajectory, the COBRA platform can adapt its movement to navigate through complex environments while maintaining stability and efficiency. The parameters \( A \), \( \omega \), and \( \phi \) in the sine equation allow for precise control over the sidewinding motion, enabling the platform to adjust its speed, direction, and curvature as needed. This flexibility is crucial for applications such as search and rescue missions, exploration of uneven terrains, and inspection tasks in challenging environments.

The ground model used in our simulations is given by 
\begin{equation}
\begin{aligned}
    F_{GRF} &= \begin{cases} \, 0 ~~  \mbox{if } p_{C,z} > 0  \\
     [F_{GRF,x},\, F_{GRF,y},\, F_{GRF,z}]^\top ~~ \mbox{else} \end{cases} \\
\end{aligned}
\label{eq:grf}
\end{equation}
\noindent where $p_{C,i},~~i=x,y,z$ are the $x-y-z$ positions of the contact point; $F_{GRF,i},~~i=x,y,z$ are the $x-y-z$ components of the ground reaction force assuming a contact takes place between the robot and the ground substrate. In Eq.~\ref{eq:grf}, the force terms are given by 
\begin{equation}
     F_{GRF,z} = -k_1 p_{C,z} - k_{2} \dot p_{C,z}, 
    \label{eq:grf-z}
\end{equation}
% \noindent and 
\begin{equation}
    F_{GRF,i} = - s_{i} F_{GRF,z} \, \mathrm{sgn}(\dot p_{C,i}) - \mu_v \dot p_{C,i} ~~  \mbox{if} ~~i=x, y
    \label{eq:grf-xy}
\end{equation}
\noindent In Eqs.~\ref{eq:grf-z} and \ref{eq:grf-xy}, $k_{1}$ and $k_{2}$ are the spring and damping coefficients of the compliant surface model. In Eq.~\ref{eq:grf-xy}, the term $s_i$ is given by 
\begin{equation}
    s_{i} = \Big(\mu_c - (\mu_c - \mu_s) \mathrm{exp} \left(-|\dot p_{C,i}|^2/v_s^2  \right) \Big)
    \label{eq:grf-s}
\end{equation}
\noindent where $\mu_c$, $\mu_s$, and $\mu_v$ are the Coulomb, static, and viscous friction coefficients; and, $v_s > 0$ is the Stribeck velocity. Specifically speaking, from Eqs.~\ref{eq:full-dynamics} and \ref{eq:grf}, the following parameters are unknown to us: actuator models parameters and Stribeck terms \cite{liljeback_simplified_2010}.

\section{Motivation}

My strategy to address the sim2real gap involves employing the Reinforcement Learning-based Model Matching approach. This method aims to identify the accurate parameters of COBRA's dynamic model within the simulation environment, thus bridging the gap between simulation and reality. By leveraging these precise parameters, we can develop and train advanced locomotion and manipulation strategies using learning-based techniques. These strategies can then be effectively transferred and implemented on the real robot, enhancing its performance and capabilities. It is crucial to highlight that the parameters referenced in Fig. \ref{fig:disparity} were determined through a human trial-and-error approach. It is imperative not to rely on simplistic linear search or extensive trial-and-error methods for model tuning. These methods are computationally demanding and unsuitable for addressing the sim2real gap across various platforms. What we need is a modular solution capable of tuning the model based on its past interactions and applicable to solving sim2real challenges across diverse robotic platforms. This is where the Model Matching Framework becomes invaluable. The Model Matching Framework is designed to tune the dynamic model and identify accurate parameters that minimize the disparity between simulation and reality using a reinforcement learning-based method.

\subsection{Markov Decision Process}

A \ac{MDP} is a mathematical framework used to model decision-making problems in environments with uncertainty and sequential actions. It is widely applied in fields like artificial intelligence, operations research, and economics to formulate and solve problems involving decision making under uncertainty.

At the core of an MDP are several key components:

\begin{itemize}
    \item \textbf{States (S):} A set of states representing all possible configurations of the environment. Denoted as \( S = \{ s_1, s_2, ..., s_n \} \), where \( n \) is the total number of states.
    
    \item \textbf{Actions (A):} A set of actions available to an agent in each state. Denoted as \( A = \{ a_1, a_2, ..., a_m \} \), where \( m \) is the total number of actions.
    
    \item \textbf{Transition Probabilities (T):} The probabilities of transitioning from one state to another after taking a specific action. Represented as \( T(s, a, s') \), where \( s \) is the current state, \( a \) is the action taken, and \( s' \) is the resulting state.
    
    \item \textbf{Rewards (R):} Immediate rewards received by the agent upon transitioning from one state to another. Denoted as \( R(s, a, s') \), where \( s \) is the current state, \( a \) is the action taken, and \( s' \) is the resulting state.
    
    \item \textbf{Policy ($\pi$):} A strategy or set of rules that guides the agent's decision-making process. It defines which action to take in each state. Denoted as \( \pi(s) \), representing the action chosen in state \( s \) according to the policy.
\end{itemize}

The dynamics of an MDP can be illustrated using a diagram called a Markov Decision Process diagram. This diagram typically consists of circles representing states, arrows indicating possible transitions between states based on actions, and associated probabilities and rewards. Mathematically, the MDP framework can be defined using the Bellman equation, which expresses the value of a state in terms of the expected immediate reward and the value of the next state:

\[ V(s) = \max_{a \in A} \left( R(s, a) + \gamma \sum_{s' \in S} T(s, a, s') \cdot V(s') \right) \]

Here, \( V(s) \) represents the value of state \( s \), \( R(s, a) \) is the immediate reward for taking action \( a \) in state \( s \), \( \gamma \) is the discount factor that determines the importance of future rewards, \( T(s, a, s') \) is the transition probability from state \( s \) to state \( s' \) after taking action \( a \), and \( \max_{a \in A} \) denotes the maximum over all possible actions.

\subsection{Reinforcement Learning}

\begin{figure}
  \centering
  \includegraphics[width=0.5\linewidth]{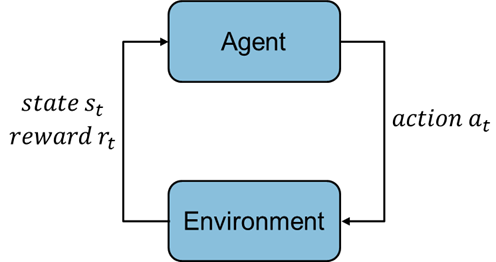} 
  \caption{Reinforcement Learning Problem Setup}
  \label{fig:rl}
\end{figure}

RL is a machine learning paradigm where an agent learns to make decisions by interacting with an environment to maximize cumulative rewards \cite{schaul_prioritized_2016} \cite{wang_sample_2017}. RL is based on the concept of trial and error, where the agent takes actions in the environment, receives feedback in the form of rewards, and adjusts its strategy over time to achieve optimal outcomes \cite{wang_dueling_2016}.

At the core of \ac{DRL} is the \ac{MDP}, which formalizes the \ac{DRL} framework. An \ac{MDP} consists of a set of states \( S \), a set of actions \( A \), transition probabilities \( T \), immediate rewards \( R \), and a policy \( \pi \). Mathematically, an \ac{MDP} can be represented as \( (S, A, T, R, \pi) \). The agent interacts with the environment by following a policy \( \pi \), which determines the action to take in each state. The goal of the agent is to learn an optimal policy \( \pi^* \) that maximizes the expected cumulative reward over time \cite{wang_dueling_2016}. The optimal policy is often denoted as \( \pi^*(s) \), where \( s \) is the current state.

The agent learns to make decisions through trial and error using the following key components:

\begin{itemize}
    \item \textbf{Value Function:} The value function estimates the long-term desirability of states or state-action pairs. There are two types of value functions: the state value function \( V(s) \), which estimates the value of being in a particular state, and the action value function \( Q(s, a) \), which estimates the value of taking a specific action in a given state.

   \[ V(s) = \mathbb{E}[R_{t+1} + \gamma R_{t+2} + \gamma^2 R_{t+3} + \dots | S_t = s] \]
   
   \[ Q(s, a) = \mathbb{E}[R_{t+1} + \gamma R_{t+2} + \gamma^2 R_{t+3} + \dots | S_t = s, A_t = a] \]
   
   Here, \( \gamma \) is the discount factor that determines the importance of future rewards.
    
    \item \textbf{Policy Evaluation and Improvement:} The agent evaluates the current policy by estimating the value of states or state-action pairs and then improves the policy based on the value estimates. This process continues iteratively until convergence to the optimal policy.
    
    \item \textbf{Exploration vs. Exploitation:} RL involves a trade-off between exploration (trying new actions to gather information) and exploitation (using known actions to maximize rewards). Strategies such as epsilon-greedy and softmax exploration are commonly used to balance exploration and exploitation.
\end{itemize}

RL algorithms, such as Q-learning, SARSA, and \ac{DQN}, implement these concepts to enable agents to learn optimal policies in complex environments. Diagrams, such as state transition diagrams and value function diagrams, can visually represent the RL process, aiding in understanding how agents learn to make decisions through interaction with the environment.

\section{Methodology}
\label{sec:model_matching_methodology}

\begin{figure}
  \centering
  \includegraphics[width=1.0\linewidth]{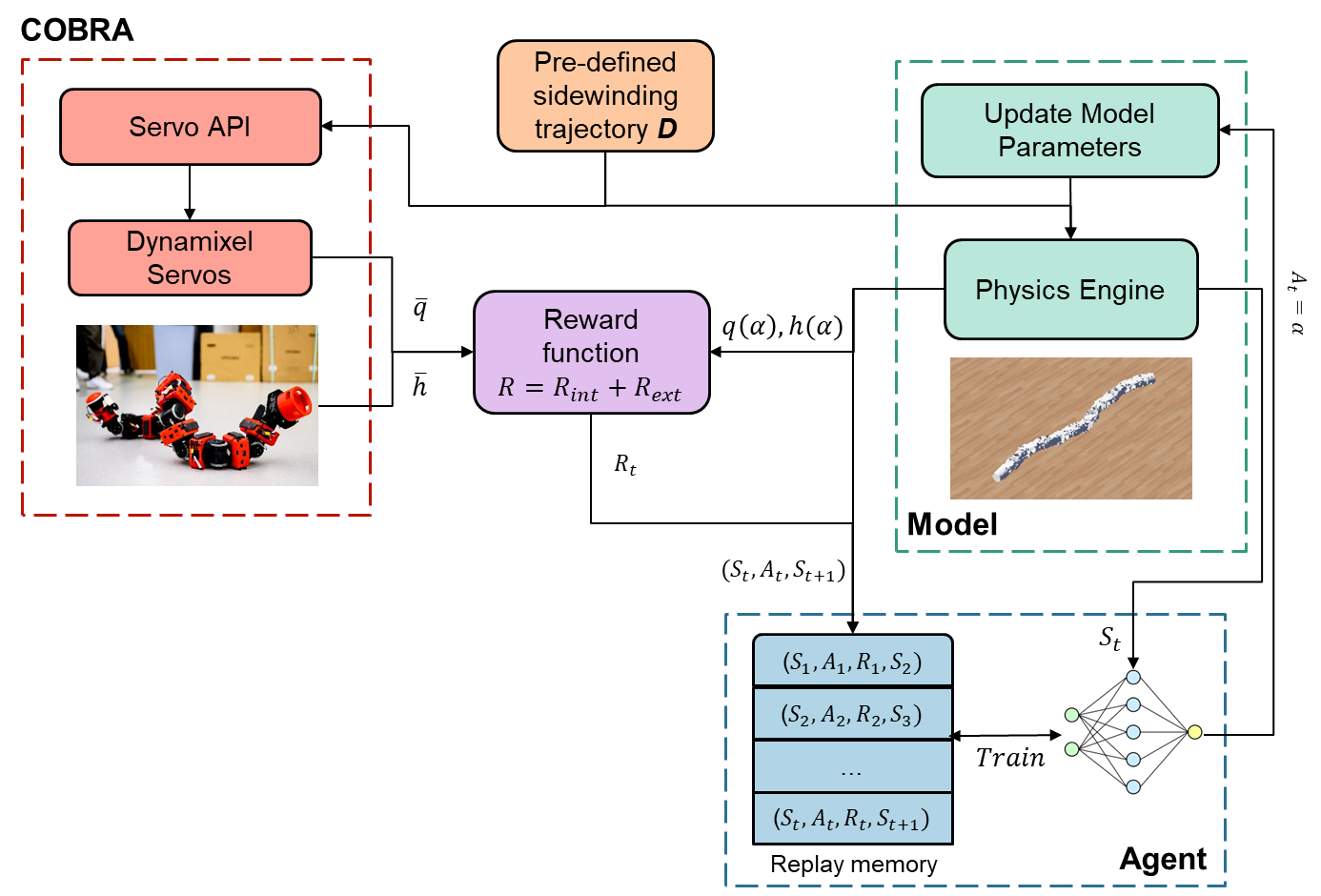} 
  \caption{Model Matching Framework}
  \label{fig:modelmatching}
\end{figure}

We approach the sim2real gap challenge by framing it as a Model Matching problem leveraging \ac{DRL}, depicted in Fig. \ref{fig:modelmatching}. Our methodology involved executing a predetermined sidewinding trajectory (operating at 0.5 Hz) denoted as \textbf{D} on both the physical robot and the Webots simulator. A sidewinding trajectory command $q_r$ is employed to guide both the real COBRA robot and its virtual model through identical movements. The performance analysis occurs in two separate domains as suggested by the partitioned model in Eq.~\ref{eq:full-dynamics}:

\begin{itemize}
    \item \textit{Underactuated dynamics:} The imposition of $q_r$ leads to the head translations $\ddot q_H$, which are underactuated. Comparing the head's position and orientation provides a clear performance metric for discerning contact force discrepancies (friction coefficients) between the physical robot and the simulation while completely excluding actuator dynamics.
    
    \item \textit{Actuated dynamics:} In addition to matching passive dynamics models, we also address actuated dynamics, specifically the joint angle trajectories $\ddot q_a$ in Eq.~\ref{eq:full-dynamics}. Aligning actuator models is crucial for evaluating the precision of the robot's movements. This comparison can be accomplished by calculating the norm distance of output joint motions between the physical robot and the simulated model.
    
\end{itemize}
Subsequently, we gathered positional data for the head, middle link, and tail, along with joint positions data throughout the sidewinding gait duration on both platforms. The alignment between simulation and reality was quantified using a reward function denoted as \textbf{R}. An RL agent was deployed to adjust the model parameters in a manner that maximizes the reward function's value, ultimately aiming to minimize the sim2real gap.

\subsection{RL Framework}

The \ac{PPO} algorithm is used to iteratively refine the model parameters. This is done with the aim of minimizing the observed discrepancies in the underactuated and actuated dynamics between the model and the actual robot. Below, we explain the state, action, and reward function defined for the PPO algorithm. I selected the PPO algorithm for addressing the sim2real gap for several reasons:

\begin{itemize}
    \item \textbf{Stability:} PPO is known for its stability during training compared to other RL algorithms like DQN or A3C. It uses a clipped objective function that prevents large policy updates, leading to more stable learning.

    \item \textbf{Sample Efficiency:} PPO tends to be more sample-efficient, meaning it requires fewer interactions with the environment to achieve good results compared to some other algorithms. This is advantageous when dealing with real-world robots where data collection can be costly and time-consuming.
    
    \item \textbf{Adaptability to Continuous Actions:} PPO naturally handles continuous action spaces, which is often the case in robotics tasks where actions like joint angles or velocities need to be controlled smoothly and precisely.
    
    \item \textbf{Balance Between Exploration and Exploitation:} PPO strikes a good balance between exploration (trying out new actions to discover better strategies) and exploitation (leveraging known good strategies). This balance is crucial for learning effective policies without getting stuck in suboptimal solutions.
    
    \item \textbf{Ease of Implementation:} PPO has a relatively simple implementation compared to some other advanced RL algorithms. It has fewer hyperparameters to tune, making it easier to experiment with and deploy in practical scenarios.
    
    \item \textbf{Robustness to Hyperparameters:} PPO is robust to changes in hyperparameters, meaning small variations in hyperparameter settings often do not lead to drastic changes in performance. This robustness simplifies the tuning process.
\end{itemize}

Overall, the combination of stability, sample efficiency, adaptability to continuous actions, balanced exploration-exploitation trade-off, ease of implementation, and robustness to hyperparameters made PPO a suitable choice for addressing the sim2real gap using RL methods.

\subsubsection{State, Action, and Reward Definition}

The state $S_t$ encapsulates the actuator parameters for internal tuning as well as Steibek terms for external tuning. The actions $A_t$ are the modifications applied to these simulation parameters.

The reward function $R = R_{\text{int}} + R_{\text{ext}}$ is given by
\begin{equation}
\begin{aligned}
    R_{\text{external}} = - \sqrt{\left( (x_{\text{des}} - x_{\text{actual}})^2 + (y_{\text{des}} - y_{\text{actual}})^2 \right)} 
\end{aligned}
\label{eq:external reward}
\end{equation}
where $x_{\text{des}}$ and $y_{\text{des}}$ denote desired x- and y-positions of the head module, respectively. $x_{\text{actual}}$ and $y_{\text{actual}}$ are the OptiTrack data. And, $R_{\text{internal}}$ is given by 
\begin{equation}
\begin{aligned}
    R_{\text{internal}} = &- (\phi_{\text{des}} - \phi_{\text{actual}})^2 - (\omega_{\text{des}} - \omega_{\text{actual}})^2  \\
    &- (A_{\text{des}} - A_{\text{actual}})^2
\end{aligned}
\label{eq:internal reward}
\end{equation}
where $\phi$, $\omega$, and $A$ are CPG variables. These reward functions are crafted to penalize the deviation in both passive dynamics and actuated dynamics. They are defined as the sum of the L$^2$-norm of the joint angle trajectory differences and the L$^2$-norm of the final head position differences.

\subsubsection{Policy Updates}

The policy search recruited here operates in a cycle of simulation runs and updates. During simulation runs, the algorithms collects data on state transitions and rewards, which are aggregated into sequences $(S_t, A_t, R_t, S_{t+1})$ and stored in a replay memory. The training phase involves updating the policy network with this data, where PPO adjusts the policy in a manner that maximizes the cumulative reward while maintaining a degree of similarity to the previous policy, using a mechanism known as clipping to avoid drastic policy changes \cite{schulman_proximal_2017}.

Through these iterative training cycles, \ac{PPO} tunes the model parameters to align the model performance with the actual COBRA robot, following a predefined sidewinding trajectory. The process continues until a convergence is reached. The following is a line-by-line breakdown of the search approach as reported in \cite{schulman_proximal_2017}:

\textit{a) Input:} The algorithm starts with initial policy parameters $\theta_0$ and initial value function parameters $\phi_0$. These parameters are what the algorithm will learn to adjust as it interacts with the environment to improve its policy.

\textit{b) Collect Trajectories:} A set of trajectories $D_k$ is collected by running the current policy $\pi(\theta_k)$ in the environment. A trajectory is a sequence of states, actions, and rewards experienced by the agent.

\textit{c) Compute rewards-to-go:} For each time step $t$, compute the total expected rewards from that time step until the end of the trajectory, denoted as $R_t$. This is used to estimate how good it is to be in a particular state.

\textit{d) Compute advantage estimates:} The advantage $A_t$ indicates how much better or worse an action is compared to the average action in a given state. This is calculated using the value function $V_{\phi}$ and the rewards-to-go.

\textit{e) Update the Policy:} First, we calculate the probability ratio, which is a fraction where the numerator is the probability of taking action $a_t$ in state $s_t$ under the new policy $\pi_{\theta}$, and the denominator is the probability of taking action $a_t$ in state $s_t$ under the old policy $\pi_{\theta_{\text{old}}}$:
\[ \text{ratio}(\theta) = \frac{\pi_{\theta}(a_t | s_t)}{\pi_{\theta_{\text{old}}}(a_t | s_t)} \]
This ratio measures how the new policy differs from the old policy in terms of the likelihood of taking the same actions.

We then clip this ratio to be within a range of $[1 - \epsilon, 1 + \epsilon]$, where $\epsilon$ is a hyperparameter typically set to a small value like 0.1 or 0.2. This clipping limits the amount by which the new policy can differ from the old one, regardless of the advantage estimate:
\[ \text{clipped}(\theta) = \text{clip}(\text{ratio}(\theta), 1 - \epsilon, 1 + \epsilon) \]
The objective function incorporates the clipped ratio and the advantage function $A_t$. It takes the minimum of the unclipped and clipped objectives, ensuring that the final objective doesn't take too large of a step (hence, "clipping"):
\[ L(\theta) = \min \left( \text{ratio}(\theta) A_t, \ \text{clipped}(\theta) A_t \right) \]
The expectation of this objective function over all timesteps and all trajectories is what the algorithm seeks to maximize. This expectation is approximated by averaging over a finite batch of timesteps and trajectories:
\[ \theta_{k+1} = \arg\max_{\theta} \left( \frac{1}{\left|D_k\right| T} \sum_{\tau \in D_k} \sum_{t=0}^{T} L(\theta) \right) \]
Finally, the parameters of the policy $\theta$ are updated using stochastic gradient ascent. This means we compute gradients of the objective function with respect to the policy parameters and adjust the parameters in the direction that increases the objective:
\[ \theta_{\text{new}} \leftarrow \theta_{\text{old}} + \alpha \nabla_{\theta} \left( \frac{1}{\left|D_k\right| T} \sum_{\tau \in D_k} \sum_{t=0}^{T} L(\theta) \right) \]
where $\alpha$ is the learning rate. The Adam optimizer is often used for this step because it adapts the learning rate for each parameter, helps in converging faster, and is more robust to the choice of hyperparameters.

By maximizing this objective, PPO-Clip seeks to improve the policy by making sure that the actions that would increase the expected return are taken more frequently, while ensuring that the policy does not change too drastically, which could lead to poor performance due to overfitting to the current batch of data. The results of the tuning procedure are presented in Chapter \ref{chap:results}.

%% locomanipualtiom
\chapter{Loco-manipulation Controller}
\label{chap:locomanipulation}

\section{Motivation}

The tuned model derived from the Model Matching framework offers valuable capabilities for developing sophisticated locomotion strategies that can be transferred to the real robot. Among these strategies is Loco-manipulation \cite{schneider_reachbot_2021} \cite{duz_robotic_2022}, a technique that involves manipulating objects in the vicinity of the robot during locomotion tasks. This aspect is particularly significant for the COBRA platform as it leverages its distinctive design to augment dexterity and precision in object manipulation scenarios \cite{salagame_loco-manipulation_2024} \cite{salagame_non-impulsive_2024}. By integrating locomotion and manipulation seamlessly, COBRA can enhance its functionality and adaptability across a range of tasks, contributing to its overall efficiency and versatility. In the subsequent sections, I introduce the Loco-manipulation controller, detailing its design and functionality. While the Loco-manipulation controller effectively addresses the specified problem, it's crucial to highlight that the current controller version cannot be directly transferred to the real robot. This limitation arises from the fact that the locomotion policy utilized in this controller was trained using an untuned model. The primary objective of this chapter is to delve into the algorithms behind the said Loco-manipulation controller.

\section{Pure Pursuit Algorithm}

\begin{figure}
  \centering
  \includegraphics[width=0.7\linewidth]{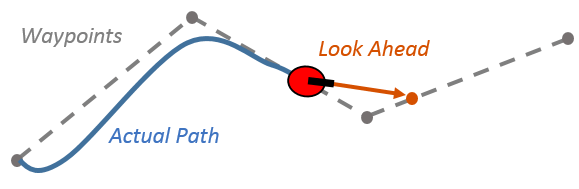} 
  \caption{Illustrates the look ahead vector and steering angle in pure pursuit algorithm \cite{samuel_control_2018}.}
  \label{fig:pure_pursuit}
\end{figure}

The loco-manipulation controller is based on a classical robotics navigation algorithm called "Pure Pursuit Algorithm". The Pure Pursuit algorithm is a path-tracking algorithm widely used in robotics and autonomous vehicle navigation. It is particularly effective for systems with differential steering, such as cars or robots with two-wheel drive systems. The main idea behind the Pure Pursuit algorithm is to guide the vehicle or robot to follow a desired path or trajectory by continuously adjusting its steering angle based on the current position and orientation relative to the path.

The algorithm operates by calculating the look-ahead point on the path that the vehicle should aim for. This look-ahead point is determined based on the vehicle's current position and orientation. A circle with a fixed radius, called the look-ahead distance \( L\), is defined around the vehicle. The look-ahead point is then selected as the point on the path that intersects with this circle, typically chosen as the point closest to the circle's circumference.

\begin{figure}
  \centering
  \includegraphics[width=0.7\linewidth]{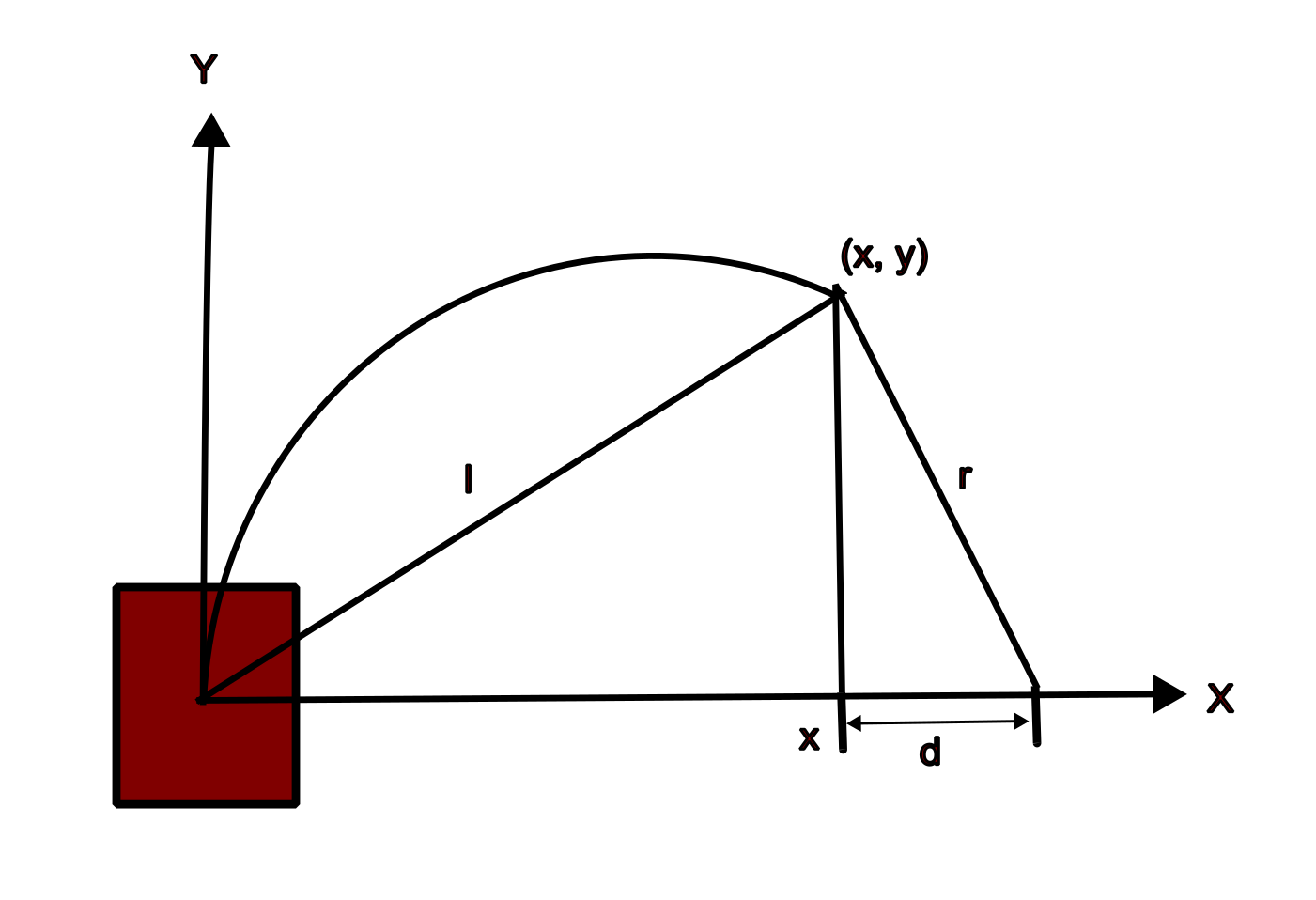} 
  \caption{Geometry of the pure pursuit algorithm.}
  \label{fig:pp_geo}
\end{figure}

Consider the illustration provided in Fig. \ref{fig:pp_geo}, depicting the vehicle along with its coordinate system axes. The vehicle's X-axis aligns with the rear axle, as detailed by Shin[2], demonstrating that the propulsion and steering mechanisms become geometrically independent when the vehicle's coordinate system is positioned at the rear differential with the x-axis aligned to the rear axle. Additionally, the diagram includes the point (x,y), located at a distance of one lookahead distance from the origin and constrained to lie on the path. The primary objective is to compute the curvature of the arc connecting the origin to (x,y) while maintaining a chord length of 1.

The subsequent pair of equations are as follows. The first equation is derived from the geometry observed in the smaller right triangle within Fig. \ref{fig:pp_geo}, while the second equation results from summing the line segments along the x-axis:

\[ x + y = I \quad (2.1) \]
\[ x + d = r \quad (2.2) \]

Equation (2.1) delineates the circle with a radius of 1 around the origin, representing the locus of possible goal points for the vehicle. On the other hand, Equation (2.2) defines the relationship between the radius of the arc connecting the origin and the goal point, along with the x offset of the goal point from the vehicle. This equation essentially states that the radius of the arc and the x offset are independent and differ by d.

Subsequent equations in the series establish the relationship between the curvature of the arc and the lookahead distance. The algebraic manipulations involved are straightforward and do not necessitate further elaboration.

\[ d = x - r \]
\[ (r - x)^2 + y^2 = r^2 \]
\[ r^2 - 2rx + x^2 + y^2 = r^2 \]
\[ 2rx = l^2 \]
\[ r = \frac{l^2}{2x} \]
\[ \gamma = \frac{2x}{l^2} \]

The curvature has been linked to the x-offset of the goal point from the origin using the inverse square of the lookahead distance. This relationship shares a similar form with a proportional controller, where the gain is twice the inverse square of the lookahead distance. However, in this context, the "error" refers to the x-offset of a point positioned ahead of the vehicle.

The implementation of the pure pursuit algorithm is relatively straightforward and can be outlined as follows:

\textbf{1) Determine the current location of the vehicle:} The vehicle's central controller provides functions that report its current position as (x, y, heading) with respect to the global reference frame established at initialization.

\textbf{2) Find the path point closest to the vehicle:} As mentioned in the geometric derivation, the goal point is within one lookahead distance from the vehicle. If there are multiple points at this distance, the vehicle steers toward the closest one.

\textbf{3) Find the goal point:} Move along the path to calculate the distance between each path point and the vehicle's current location in global coordinates.

\textbf{4) Transform the goal point to vehicle coordinates:} Convert the goal point from global coordinates to the vehicle's local coordinates.

\textbf{5) Calculate the curvature:} Use the curvature equation derived earlier to calculate the desired curvature for the vehicle, which is then translated into steering wheel angle by the vehicle's controller.

\textbf{6) Update the vehicle's position:} During simulation or real-time operation, update the vehicle's position based on its calculated trajectory and curvature commands.

\section{Central Pattern Generators and Kuramoto Model}

\subsection{Central Pattern Generators}

\ac{CPG} are neural networks found in the spinal cord and brainstem of vertebrates, including humans. They are responsible for generating rhythmic patterns of neural activity that coordinate and control repetitive motor movements, such as walking, swimming, and breathing, without requiring continuous input from higher brain centers. \ac{CPG} play a crucial role in generating and controlling locomotion and other rhythmic behaviors, making them a fundamental component of the neural circuitry underlying motor control.

\begin{figure}
  \centering
  \includegraphics[width=0.7\linewidth]{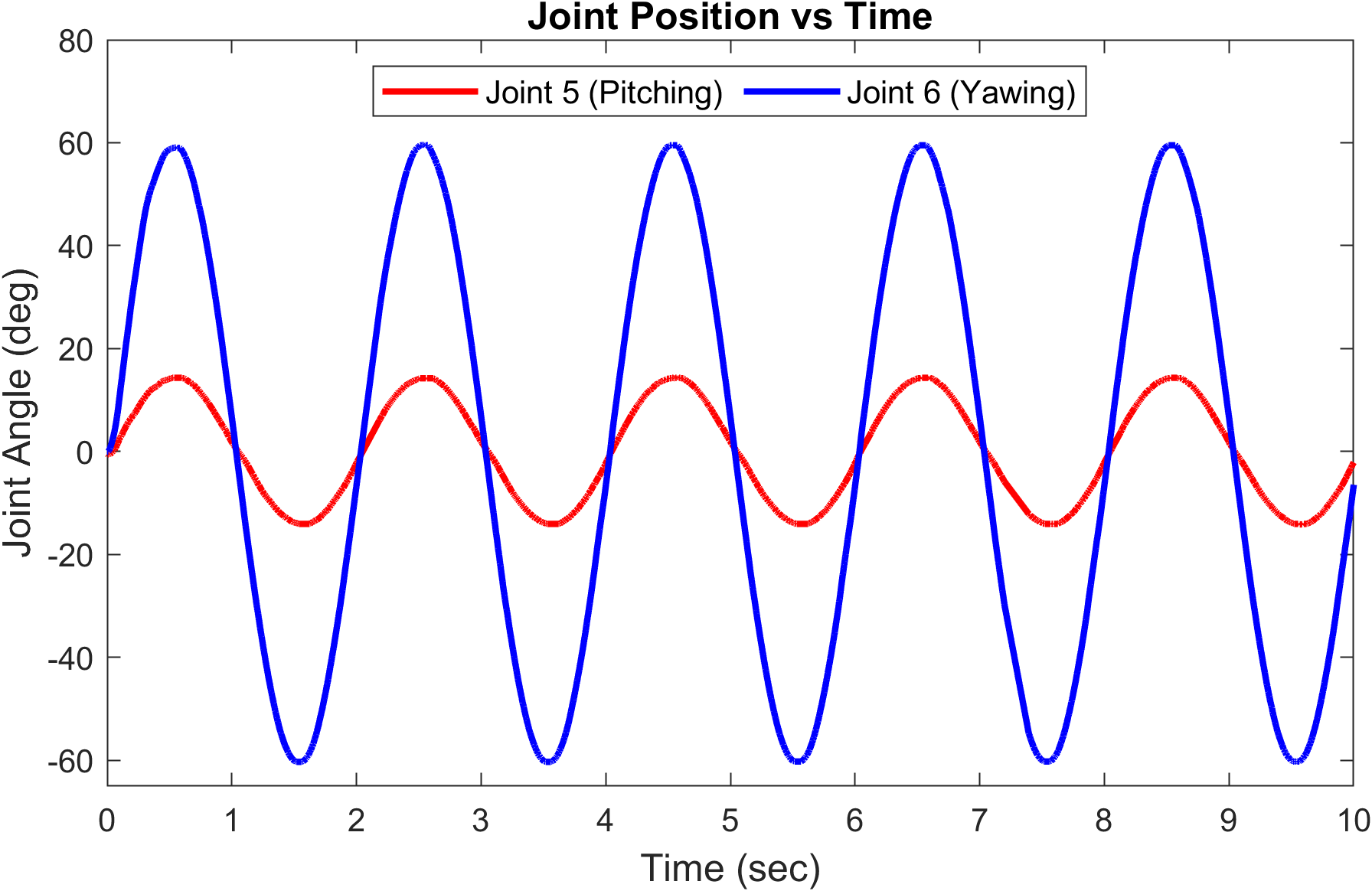} 
  \caption{Signals generated from a simple sine wave CPG.}
  \label{fig:cpg_eg}
\end{figure}

One key characteristic of CPGs is their ability to produce rhythmic patterns of neural activity even in the absence of sensory feedback or external stimuli. This autonomous rhythmicity allows organisms to perform rhythmic movements automatically and adaptively, without the need for constant conscious control. For example, during walking, CPGs generate the alternating patterns of muscle activation and relaxation in the legs that produce the coordinated movement of stepping forward. Figure \ref{fig:cpg_eg} represents typical wave forms of the spiking-bursting behaviour generated by the Rulkov map for two different sets of parameters

CPGs operate through complex interactions between excitatory and inhibitory neurons within the network. These interactions give rise to oscillatory patterns of neural activity, with specific groups of neurons firing in coordinated sequences to produce rhythmic motor output. The exact organization and connectivity of CPGs can vary across species and types of movements but generally involve interconnected neural circuits that regulate the timing and coordination of muscle contractions.

In research and robotics applications, CPGs have been studied extensively due to their potential for biomimetic control of locomotion and other rhythmic behaviors in artificial systems. By modeling the principles of CPGs in artificial neural networks, researchers can develop control systems that mimic the robust and adaptive rhythmicity seen in biological organisms. These artificial CPGs can then be integrated into robotic systems to control locomotion, navigation, and other motor functions autonomously and efficiently.

Furthermore, studying CPGs provides insights into the underlying mechanisms of motor control and coordination in living organisms. Understanding how CPGs generate and regulate rhythmic patterns of activity can inform research in neuroscience, rehabilitation engineering, and robotics, leading to advancements in prosthetics, assistive devices, and autonomous robotic systems capable of complex locomotion and manipulation tasks.

\subsection{Modified Kuramoto Model}

The Kuramoto CPG model is a mathematical framework used to describe the rhythmic oscillatory behavior observed in biological systems, particularly in the context of locomotion control in robots. It is based on the Kuramoto model from physics, which represents a network of coupled phase oscillators. In the CPG context, each oscillator corresponds to a motor unit or joint, and their interactions generate rhythmic patterns resembling biological locomotion. The Kuramoto CPG model allows for the synchronization and coordination of these oscillators, enabling the generation of complex locomotion patterns such as walking, crawling, or swimming in robotic systems. We used a modified version of the original Kuramoto Model. The dynamics of Modified Kuramoto CPG Model are shown below:

\[
\dot{\varphi} = \omega + A \cdot \varphi + B \cdot \theta
\]

\[
\ddot{r} = a \cdot [\frac{a}{4} \cdot (R - r) - \dot{r}]
\]

\[
x = r \cdot \sin(\varphi) + \delta
\]

\[
A =
\begin{bmatrix}
-\mu_1 & \mu_1 \\
\mu_2 & -2\mu_2 & \mu_2 \\
 &  & \ddots &  \\
 &  & \mu_{n-1} & -2\mu_{n-1} & \mu_{n-1} \\
&  & & \mu_n & -\mu_n \\
\end{bmatrix}
\]

\[
B =
\begin{bmatrix}
1 \\
-1 & 1 \\
& -1 & \ddots \\
& & \ddots & 1 \\
& & & -1 & 1 \\
& & & & -1 \\
\end{bmatrix}
\]

Where \(\varphi \in \mathbb{R}^n\) and \(r \in \mathbb{R}^n\) are the internal states of CPG, \(n\) is the number of output channels (usually equal to the number of robot joints), \(a\) and \(\mu_i\) are hyperparameters controlling the convergence rate, \(R \in \mathbb{R}^n\), \(\omega \in \mathbb{R}^n\), \(\theta \in \mathbb{R}^{n-1}\), \(\delta \in \mathbb{R}^n\) are inputs controlling the desired amplitude, frequency, phase shift, and offset, and \(x \in \mathbb{R}^n\) is the output sinusoidal waves of \(n\) channels.

\section{Locomotion Policy}

\begin{figure}
  \centering
  \includegraphics[width=0.9\linewidth]{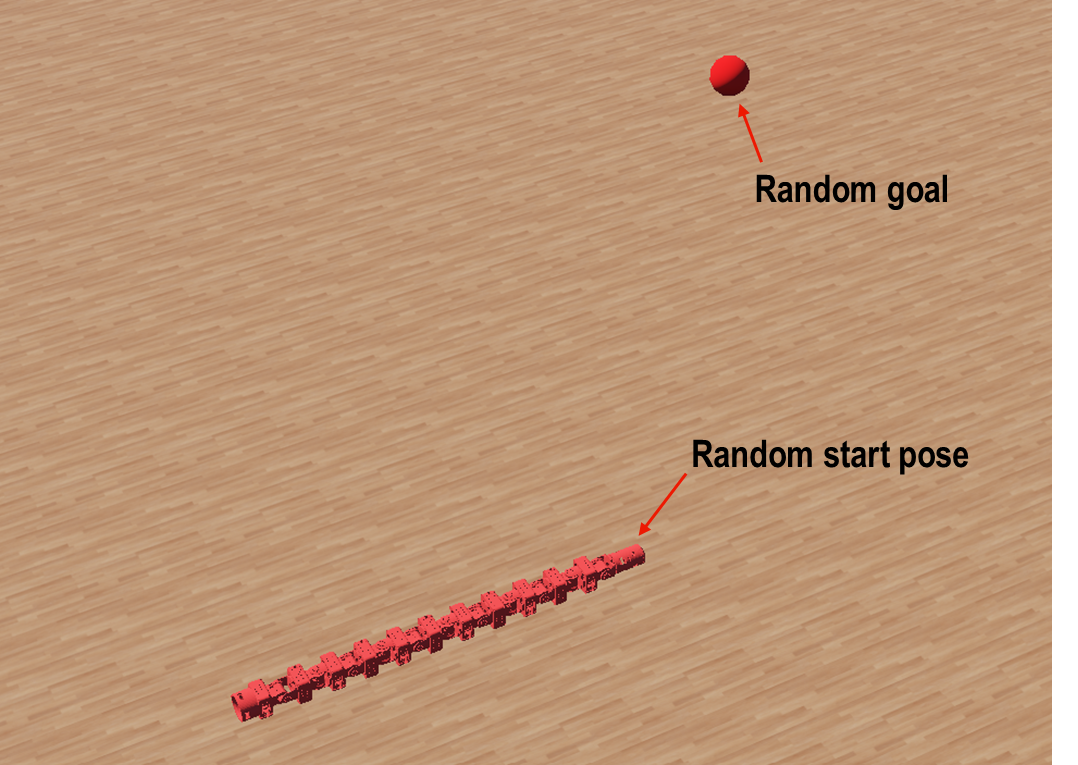} 
  \caption{Snapshot of Locomotion Policy Setup in Simulator}
  \label{fig:loco_policy}
\end{figure}

The loco-manipulation controller is derived from a pre-existing locomotion policy, which was trained using an untuned simulator model  \cite{jiang_hierarchical_2023} \cite{jiang_hierarchical_2023}. It's crucial to note that this controller cannot be directly applied to the real robot at this stage. A detailed explanation of the training process for this locomotion policy is provided in the subsequent paragraphs.

Agent training was conducted across a varied range of randomly generated curriculum terrains. In each agent, a \ac{CPG} module is incorporated, and the parameters of the corresponding \ac{CPG} module are fine-tuned by the actor's outputs. Through this \ac{CPG} parameter adjustment process, optimized gaits suitable for their designated curriculum terrains are developed by agents. The training landscapes, created using Perlin noise with dimensions of 16m × 16m, prompt the robot to navigate and reach a random goal pose from any starting location in each training episode. The associated \ac{MDP} is delineated in the following manner \cite{sutton_policy_1999} \cite{fujimoto_addressing_2018}.

The state space encompasses the robot's state components and tactile sensor readings. This includes joint positions \( \mathbb{R}^n \), IMU readings \( \mathbb{R}^3 \), spatial translation between robot and goal frames \( \mathbb{R}^3 \), and relative rotation parameterized by axis-angle system \( \mathbb{R}^4 \), totaling 21 dimensions. Utilizing only ego-centric observations from the robot streamlines the system's practicality in outdoor environments, eliminating the need for a motion capture setup as seen in previous works. The action space dictates the CPG parameters, encompassing desired amplitude \( \mathbb{R} \), frequency \( \omega \), phase shift \( \theta \), and offset \( \delta \). The reward function incentivizes the robot to reach the goal swiftly. It comprises two components: \( r_1 = \frac{1}{0.1 + d_t} \) encourages proximity to the goal, while \( r_2 = d_t^{-1} - d_t \) promotes higher velocities. These terms complement each other, with \( r_1 \) approaching 0 when the robot is distant from the goal and \( r_2 \) nearing 0 when the robot is close to the goal. Our system utilizes \ac{SAC} as the core reinforcement learning algorithm.

\section{Methodology}

\begin{figure}
  \centering
  \includegraphics[width=1.0\linewidth]{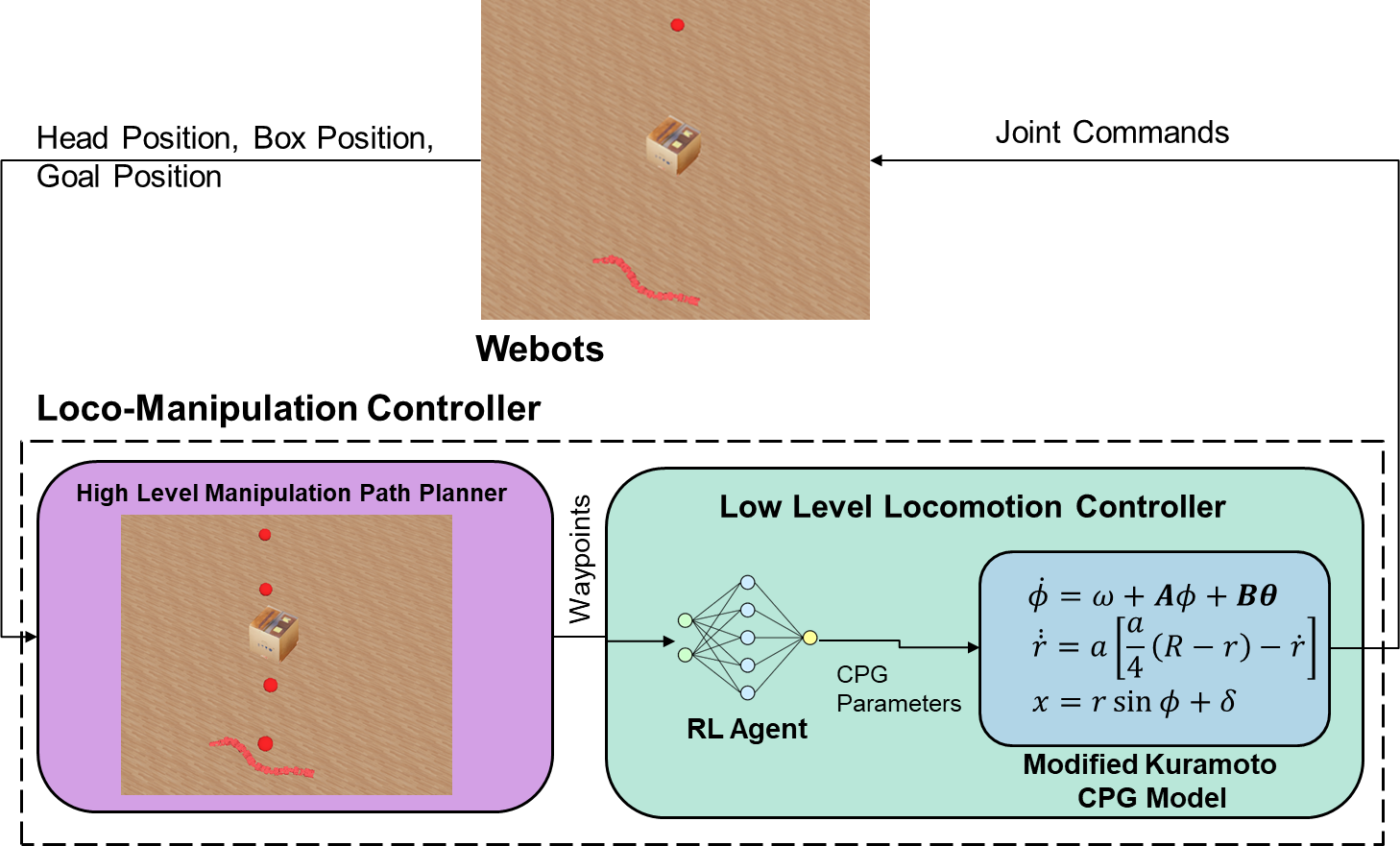} 
  \caption{Loco-manipulation Controller}
  \label{fig:locomanipulation}
\end{figure}

Figure \ref{fig:locomanipulation} shows the hierarchical framework used for developing the loco-manipulation controller. The hierarchical structure helps in breaking down the problem in two parts and it is explained step-by-step in Algorithm \ref{algo:locomanipulation}. 

\begin{algorithm}
\caption{Modified Pure Pursuit Algorithm for Loco-manipulation}
\label{algo:locomanipulation}
\begin{algorithmic}[1]
\State \textbf{Given:} head coordinates $(head_x, head_y)$, box coordinates $(box_x, box_y)$, goal coordinates $(goal_x, goal_y)$
\State \textbf{Initialize:} Thresholds $\delta_{hb}$ and $\delta_{bg}$, step length $L$, waypoints count $n$, width $W$

\Procedure{NavigateToTarget}{}
    \State \textbf{Step 1:} Select the target based on distance
    \If {$L2(\text{head}, \text{box}) < \delta_{hb}$}
        \State $target \gets box$
    \Else
        \State $target \gets goal$
    \EndIf

    \State \textbf{Step 2:} Generate $n$ waypoints between head and target
    \For {$i \gets 1$ to $n$}
        \State $x_i \gets head_x + \frac{i}{n+1}(target_x - head_x)$
        \State $y_i \gets head_y + \frac{i}{n+1}(target_y - head_y)$
    \EndFor

    \State \textbf{Step 3:} Take one step in the direction of the look ahead vector using the RL-based locomotion policy
    \State $\vec{P}_{lookahead} \gets \vec{P}_{current} + \frac{L \cdot \vec{H}}{\|\vec{H}\|}$
    \State $\alpha \gets \arctan{\left(\frac{2LW}{L2(\text{head}, \text{target})}\right)}$

    \State \textbf{Step 4:} Repeat until the box is close to the goal
    \While {$L2(\text{box}, \text{goal}) \geq \delta_{bg}$}
        \State Go back to \textbf{Step 1}
    \EndWhile
\EndProcedure
\end{algorithmic}
\end{algorithm}

The high-level manipulation path planner described in Fig. \ref{fig:locomanipulation} (the first two steps in Algorithm \ref{algo:locomanipulation}) is responsible for planning the trajectory from the robot's head position to designated goal positions. This trajectory is designed such that the robot's body can effectively manipulate an object, moving it from its initial location to the specified goal position. The manipulation planner creates waypoints along this path and transmits them to the low-level locomotion controller detailed in Fig. \ref{fig:locomanipulation} (steps 3 and 4 in Algorithm \ref{algo:locomanipulation}). The role of the low-level locomotion controller is to receive these waypoints and execute them using the locomotion policy that has been trained using Reinforcement Learning. This locomotion policy generates \ac{CPG} parameters for the Modified Kuramoto CPG Model, which in turn determines the joint positions necessary for the robot's movements. The results and analysis for the Loco-manipulation controller are presented in Chapter \ref{chap:results}.

%% results
\chapter{Results}
\label{chap:results}

This chapter showcases the results obtained through both the Model Matching framework and the Loco-Manipulation controller. Subsequent sections delve into the specific results stemming from my contributions.

\section{Model Matching}

\begin{figure}
  \centering
  \includegraphics[width=0.8\linewidth]{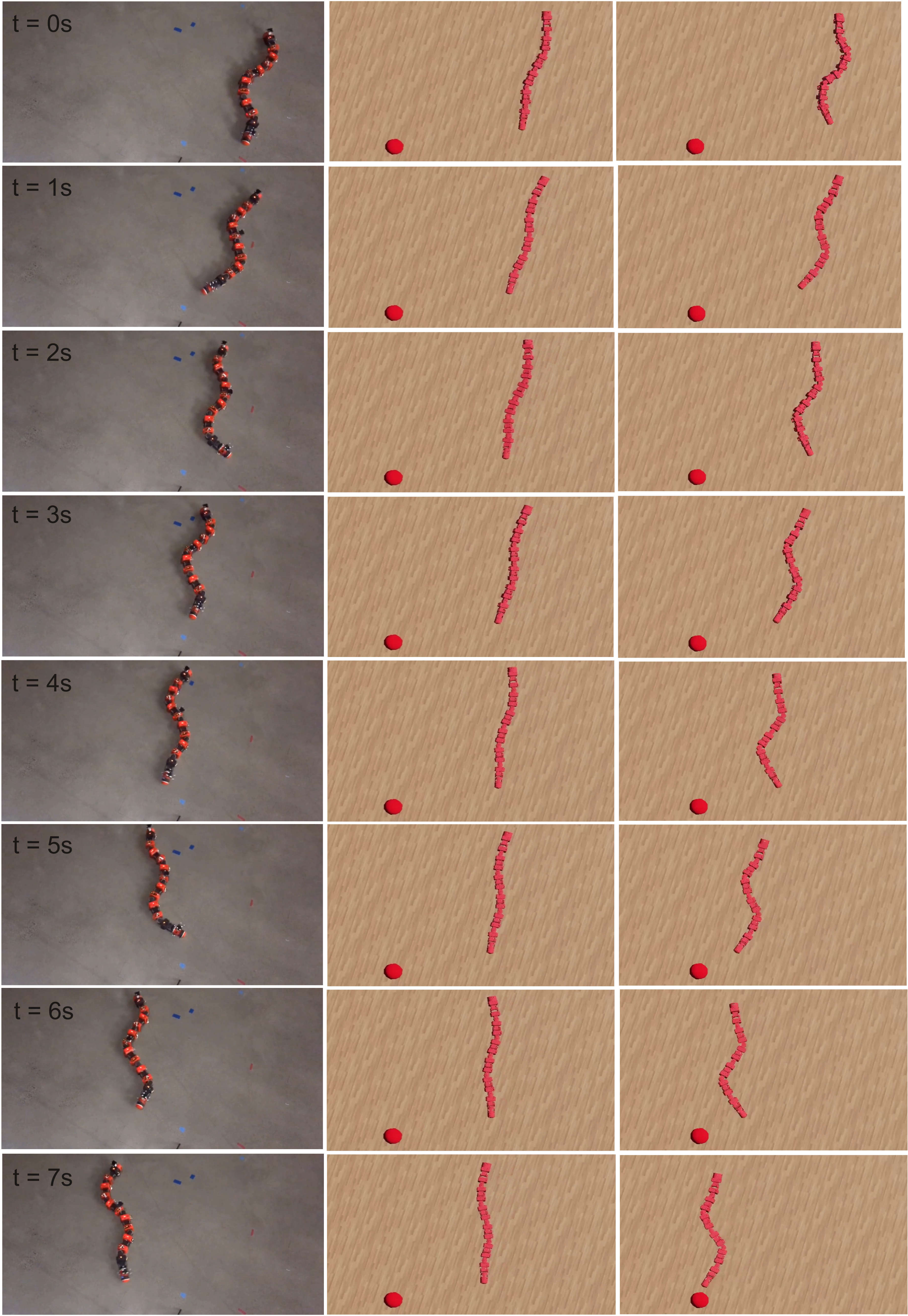} 
  \caption{Shows COBRA in Webots simulator with tuned model (right) and untuned model (center) performing the sidewinding motion. The red ball shows the location the actual robot achieved when similar joint trajectories were applied.}
  \label{fig:tuned_webots_sidewinding}
\end{figure}

In Section \ref{sec:model_matching_methodology}, the Model Matching framework details our strategy of leveraging an RL agent to finely adjust the dynamic model parameters within the simulator. Illustrating this approach, Fig. \ref{fig:tuned_webots_sidewinding} provides a visual representation of the robot executing a sidewinding gait at a frequency of 0.5 Hz, showcasing the effect of tuning the model. Notably, the red dot denotes the stopping position of the real robot post-gait execution. A direct comparison with the previous image (Fig. \ref{fig:disparity}) reveals a substantial enhancement in gait execution, signaling a step in the right direction towards addressing the sim2real problem.

\begin{figure}
  \centering
  \includegraphics[width=1.0\linewidth]{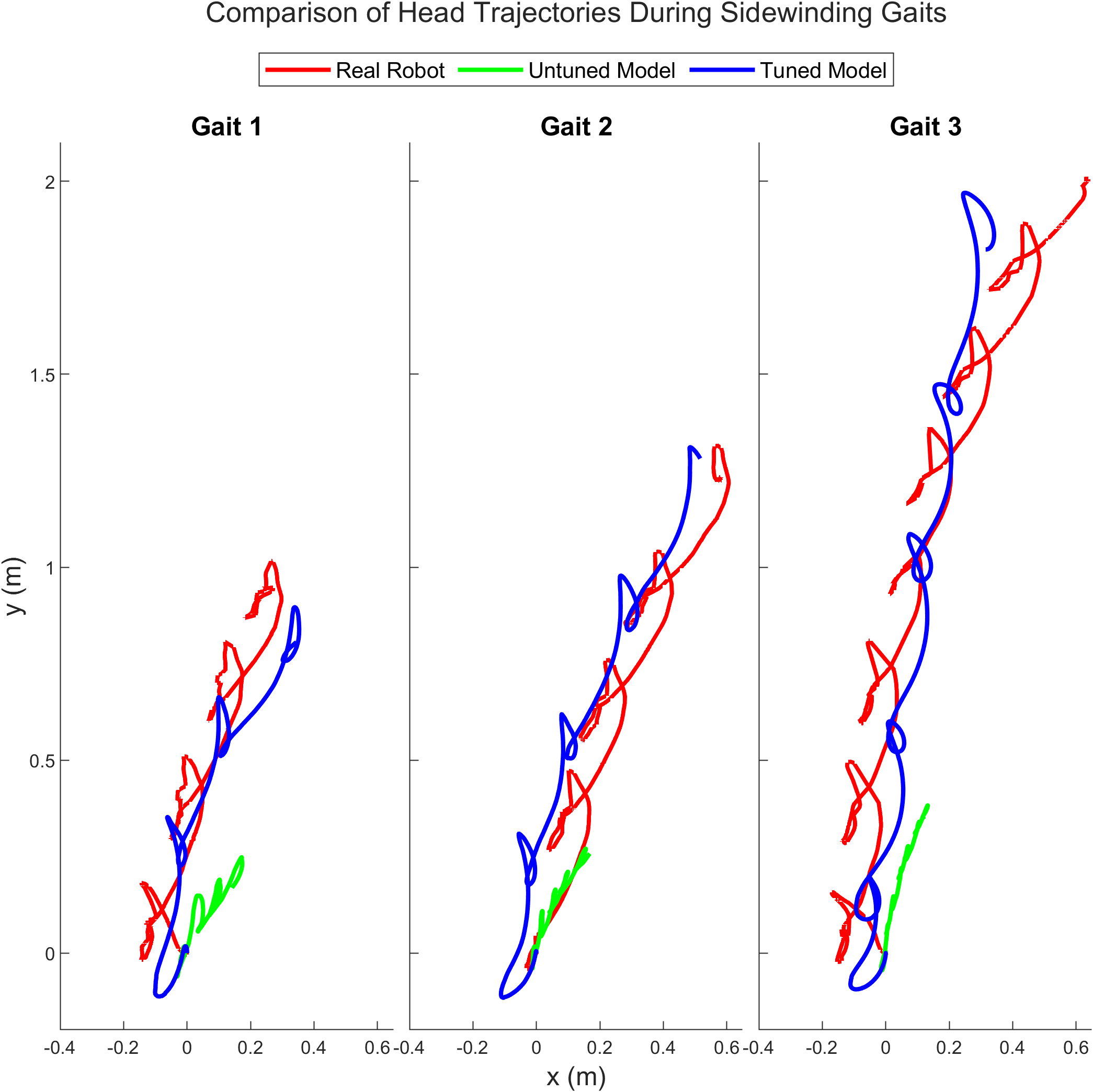} 
  \caption{Illustrates a 2D comparison between the head positions in the actual hardware platform (red), tuned model (blue) and untuned model (green) for a sidewinding trajectory \@ 0.35, 0.5, and 0.65 Hz.}
  \label{fig:head_trajectories_2d}
\end{figure}

\begin{figure}
  \centering
  \includegraphics[width=1.0\linewidth]{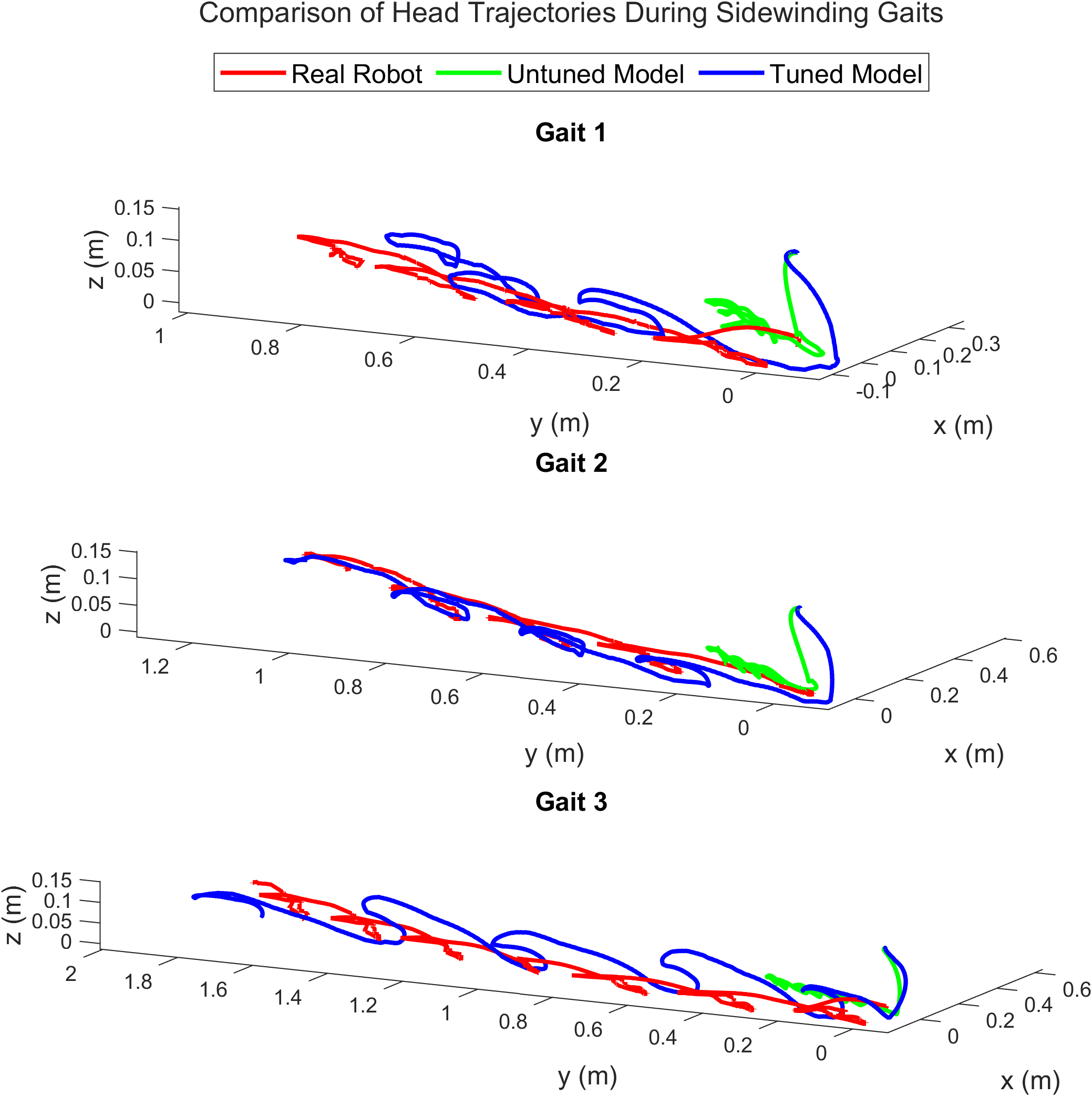} 
  \caption{Illustrates a 3D comparison between the head positions in the actual hardware platform (red), tuned model (blue) and untuned model (green) for a sidewinding trajectory \@ 0.35, 0.5, and 0.65 Hz.}
  \label{fig:head_trajectories_3d}
\end{figure}

Figure \ref{fig:head_trajectories_2d} presents a comparative analysis in 2D of three gaits—sidewinding gaits at frequencies of 0.35 Hz (gait 1), 0.5 Hz (gait 2), and 0.65 Hz (gait 3)—executed on both the real robot and on models that are untuned and tuned. It's worth noting that I exclusively trained the framework on gait 2 for parameter tuning, resulting in a noticeable alignment between the trajectory of the real robot and the tuned model, in contrast to the disparity seen with the untuned model. However, this alignment is also evident in the execution of gaits 1 and 3, showcasing the model's ability to generalize effectively through the model matching framework. Figure \ref{fig:head_trajectories_3d} shows the trajectoy comparsion in 3D for all three gaits.

\begin{figure}
  \centering
  \includegraphics[width=1.0\linewidth]{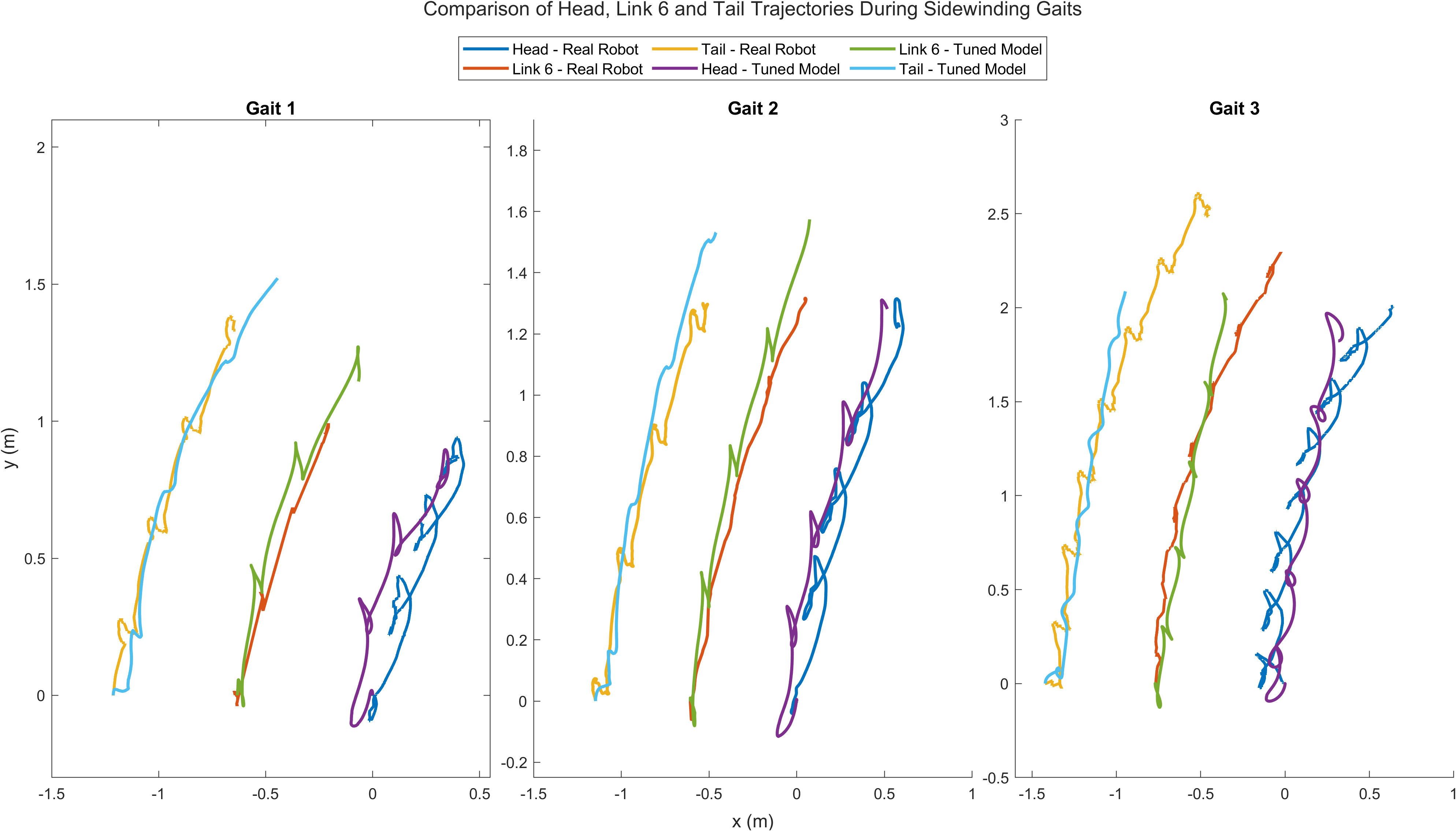} 
  \caption{Illustrates a 2D comparison between the head, middle link and tail positions in the actual hardware platform and tuned model for a sidewinding trajectory \@ 0.35, 0.5, and 0.65 Hz.}
  \label{fig:head_link_tail_trajectories_2d}
\end{figure}

\begin{figure}
  \centering
  \includegraphics[width=0.9\linewidth]{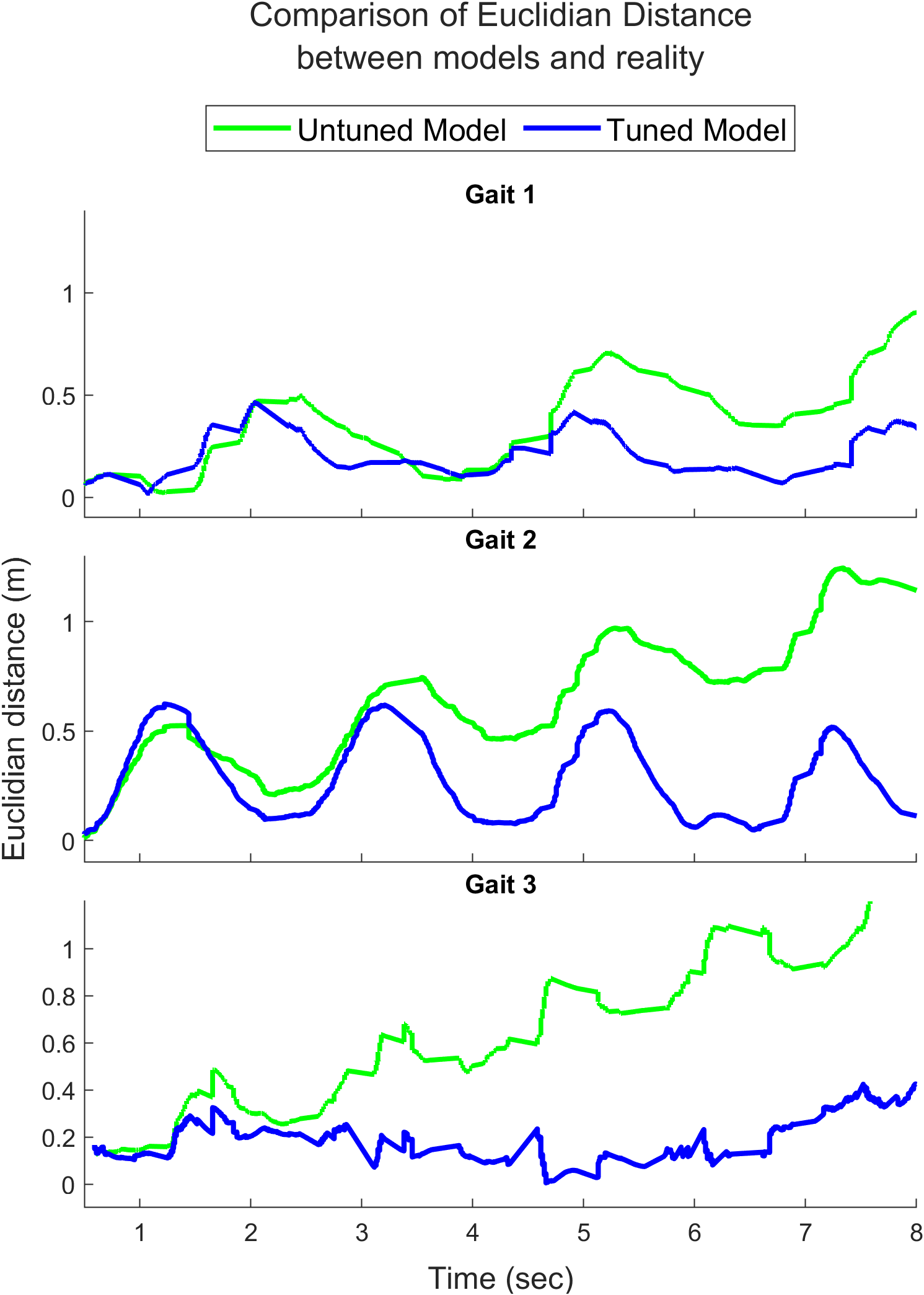} 
  \caption{Comparison of the Euclidean distance (error metric)  between the actual platform's head position captured by OptiTrack and tuned/untuned models for sidewinding @ 0.35, 0.5, and 0.65 Hz.}
  \label{fig:euclidean_distance}
\end{figure}

In Fig. \ref{fig:head_link_tail_trajectories_2d}, the alignment is also evident in the trajectories of the middle (sixth link of the robot's body) and tail links while executing the three gaits. Figure \ref{fig:euclidean_distance} presents a quantitative comparison between the tuned and untuned models using the Euclidean distance metric. This metric measures the distance between the real robot's head module position and the robot's head module position in the simulator during gait execution. The reduction in error is notable after fine-tuning the model through the Model Matching framework, particularly noticeable in gait 2 and gait 3. Given that gait 1 operates at a lower frequency of 0.35 Hz, resulting in minimal robot movement, the performance of the untuned model is comparable to that of the tuned model.

\begin{figure}
  \centering
  \includegraphics[width=1.0\linewidth]{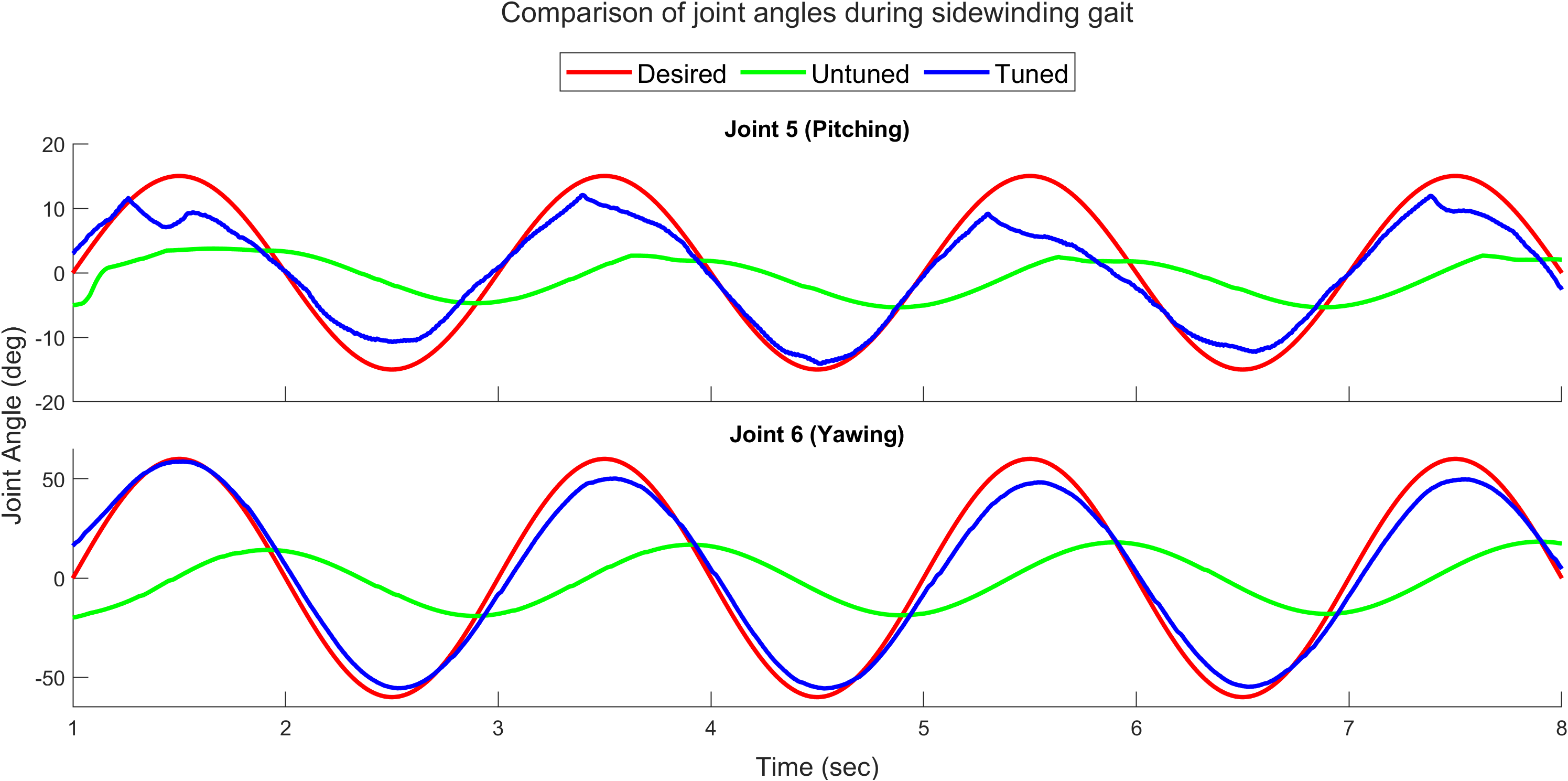} 
  \caption{Shows a comparison between the actuator joint responses from the actual hardware platform (red), the tuned (blue) and untuned (magenta) models for a sidewinding gait at @ 0.5 Hz.}
  \label{fig:joint_angle_comparison}
\end{figure}

\begin{figure}
  \centering
  \includegraphics[width=1.0\linewidth]{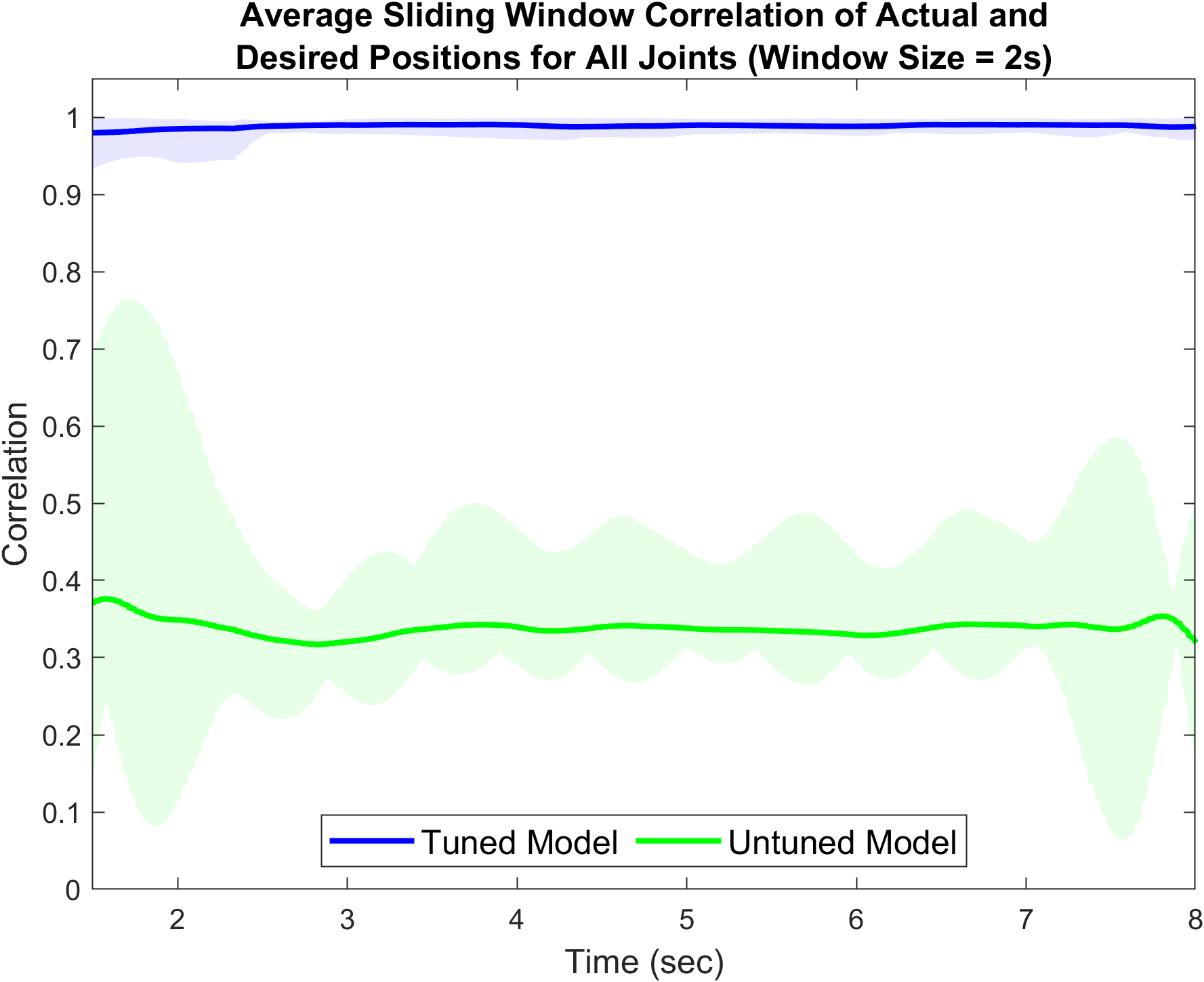} 
  \caption{Shows a comparison of average sliding window correlation (obtained by taking average of all the 11 correlation curves displayed in the above plot) of yawing joint positions for tuned and default simulators with desired signal (for the sidewinding gait @ 0.5 Hz). The model matching improved the actuator model in the the simulator as seen from the improved average correlation factor between desired and actual joint positions. Shaded regions represent the range between min and max correlation across all joints.}
  \label{fig:correlation}
\end{figure}

Figure \ref{fig:joint_angle_comparison} illustrates the joint angle comparison for joints 5 and 6 during the execution of the sidewinding gait at a frequency of 0.5 Hz. Notably, there's a noticeable improvement in agreement between the real robot and the simulator model post-tuning with the Model Matching framework. This improvement is further substantiated by Fig. \ref{fig:correlation}, which presents the comparison of average sliding window correlation across all 11 joints of the robot in the simulator and the real robot, considering both tuned and untuned models. The tuned model exhibits a correlation factor close to 1.0, indicating a strong alignment, whereas the untuned model's correlation factor hovers around 0.35, showcasing a significant disparity in performance.

\begin{figure}
  \centering
  \includegraphics[width=1.0\linewidth]{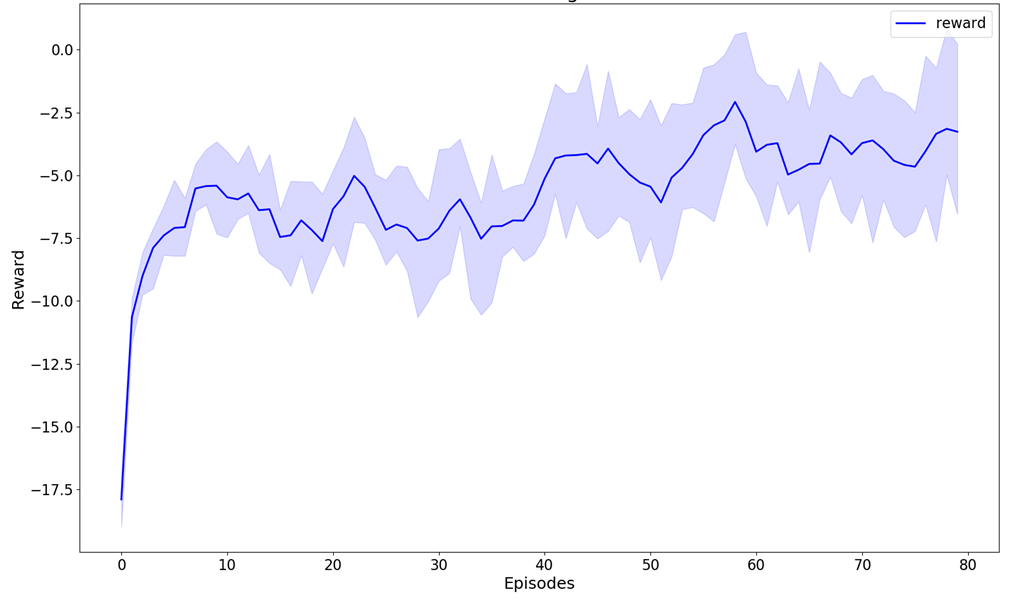} 
  \caption{Shows the training reward curve for tuning the simulator model.}
  \label{fig:reward_curve}
\end{figure}

\section{Loco-Manipulation Controller}

% \begin{figure}[htbp]
%     \centering
%     \begin{subfigure}[b]{0.7\textwidth}
%         \centering
%         \includegraphics[width=\textwidth]{fig/loco_ex1.png}
%         \caption{Caption for Image 1}
%         \label{fig:demo1}
%     \end{subfigure}
%     \hspace{0.05\textwidth}
%     \begin{subfigure}[b]{0.7\textwidth}
%         \centering
%         \includegraphics[width=\textwidth]{fig/loco_ex2.png}
%         \caption{Caption for Image 2}
%         \label{fig:demo2}
%     \end{subfigure}
%     \hspace{0.05\textwidth}
%     \begin{subfigure}[b]{0.7\textwidth}
%         \centering
%         \includegraphics[width=\textwidth]{fig/loco_ex3.png}
%         \caption{Caption for Image 3}
%         \label{fig:demo3}
%     \end{subfigure}
%     \caption{Common Caption for Both Images}
%     \label{fig:combined}
% \end{figure}

\begin{figure}
    \centering
    \includegraphics[width=1.0\linewidth]{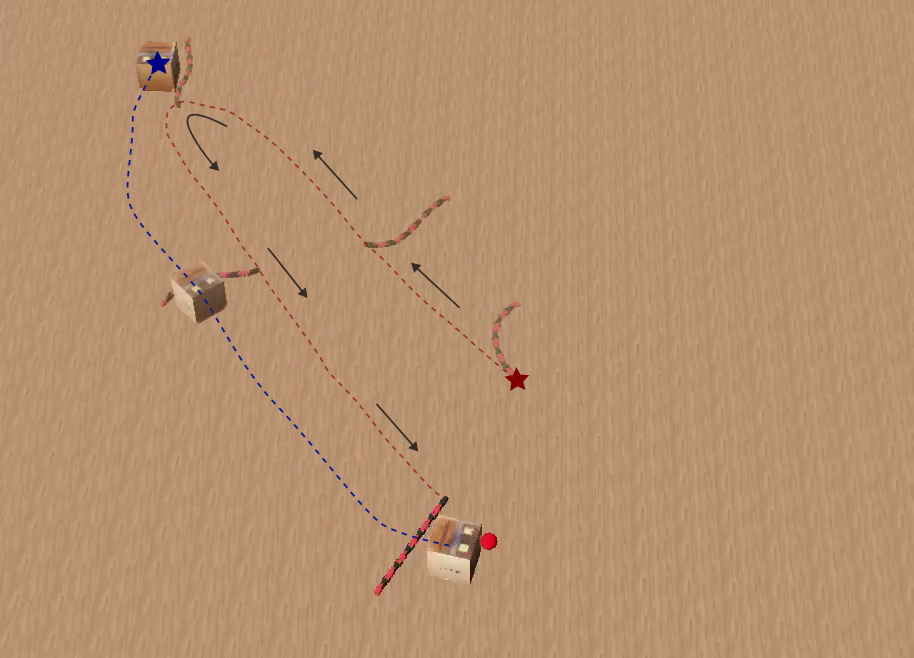}
    \caption{Shows the algorithm performing loco-manipulation inside webots simulator. Red dot is the final goal position, red star is robot's initial position, blue star is object's initial position.}
    \label{fig:demo1}
\end{figure}

\begin{figure}
    \centering
    \includegraphics[width=1.0\linewidth]{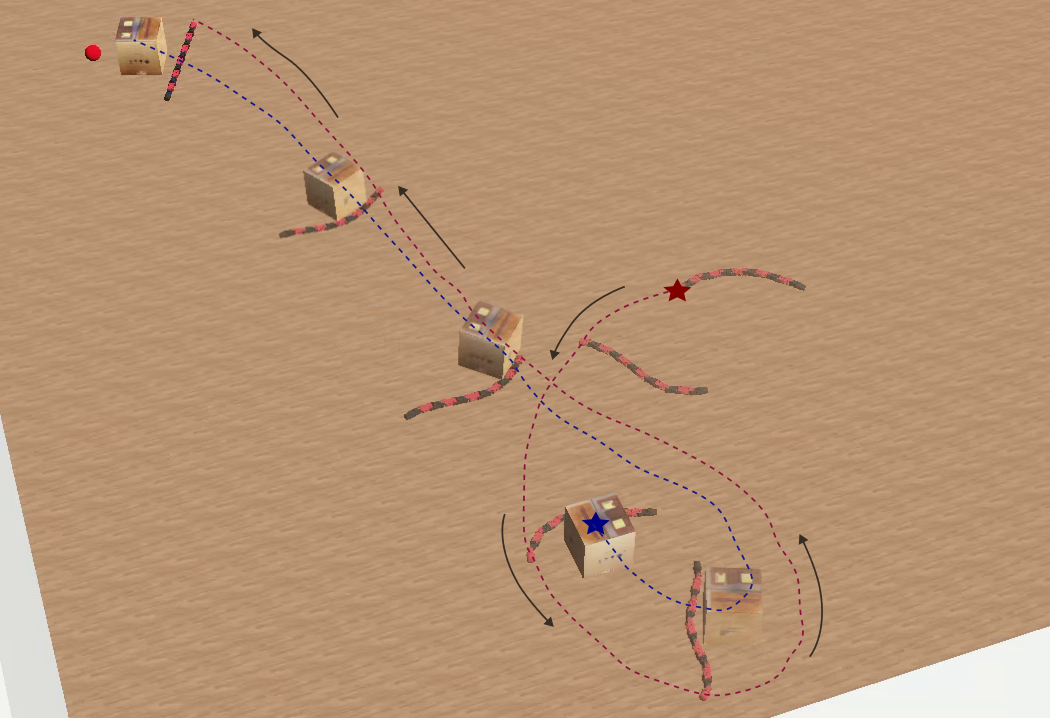}
    \caption{Shows the algorithm performing loco-manipulation with different initial robot and object positions inside webots simulator.}
    \label{fig:demo2}
\end{figure}

\begin{figure}
    \centering
    \includegraphics[width=1.0\linewidth]{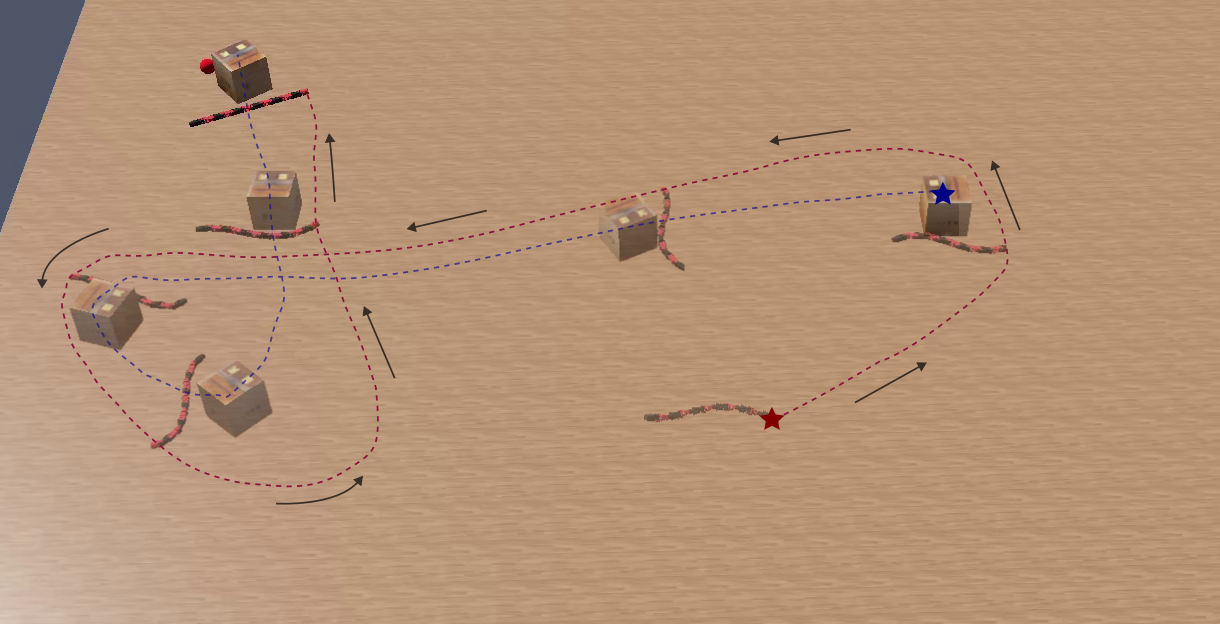}
    \caption{Shows the algorithm performing loco-manipulation with different initial robot and object positions inside webots simulator. When the object is not aligned with the goal position, the robot adjusts its course by turning while moving the object. This realignment ensures that the object moves in a straight line towards the goal position.}
    \label{fig:demo3}
\end{figure}

\begin{figure}
  \centering
  \includegraphics[width=1.0\linewidth]{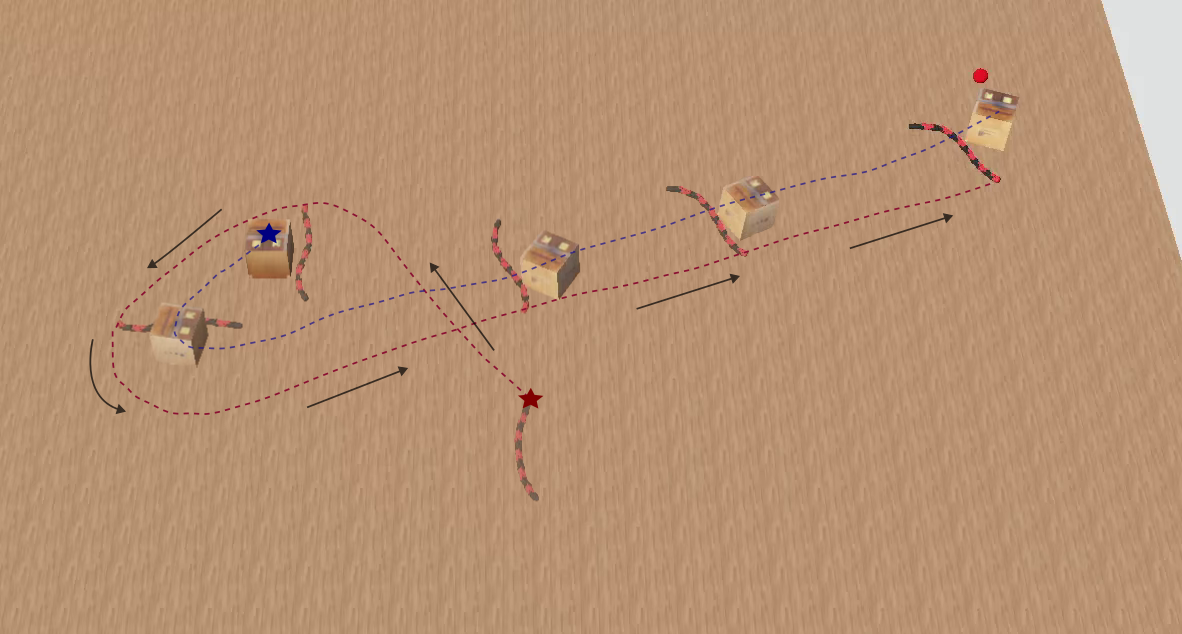} 
  \caption{Shows dynamic replanning capabilites of the algorithm. The object is relocated within the Webots simulator, shifting from its current position to a randomly chosen spot.}
  \label{fig:demo4_1}
\end{figure}

\begin{figure}
  \centering
  \includegraphics[width=1.0\linewidth]{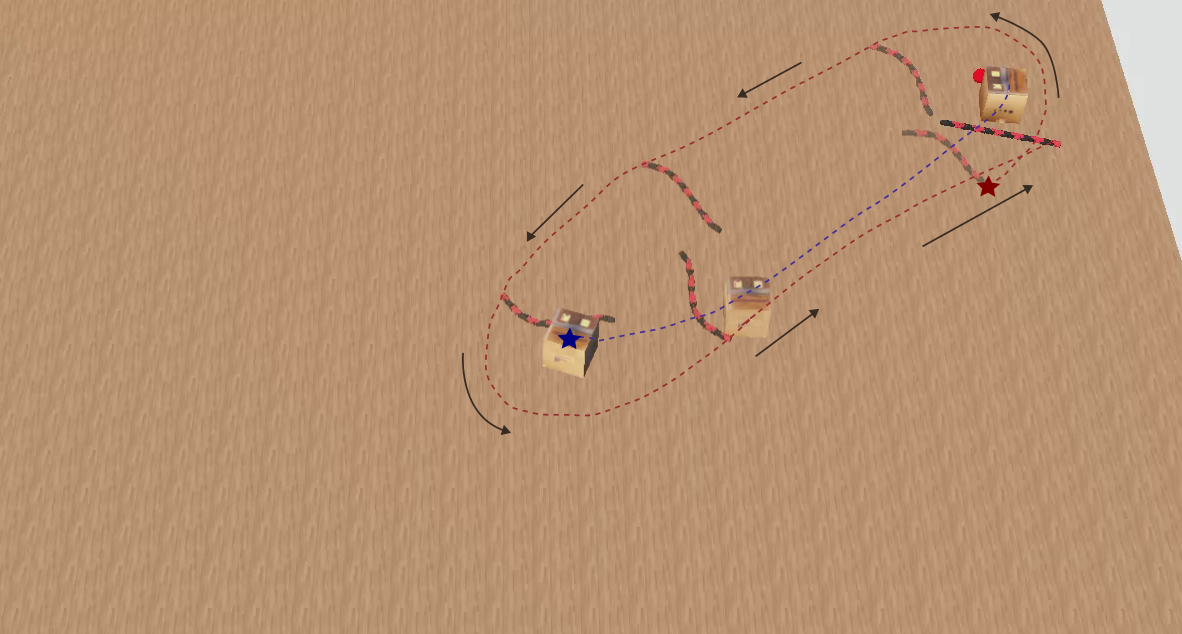} 
  \caption{Shows dynamic replanning capabilites of the algorithm. The robot effectively recalculates its path to reach the relocated object and then proceeds to move it towards the desired goal position.}
  \label{fig:demo4_2}
\end{figure}

Figure \ref{fig:demo1} depicts the successful execution of the proposed algorithm in performing loco-manipulation within the Webots simulator. Both the robot and object are initially placed at random positions in the simulated environment, with the algorithm tasked to move the object from its starting point to a predetermined final goal position. Figures \ref{fig:demo2} and \ref{fig:demo3} showcase instances where the proposed algorithm conducts loco-manipulation with varying configurations of robot, object, and goal positions. In cases where the object veers off the alignment with the goal position, the robot adjusts its trajectory by executing turns while simultaneously moving the object. This corrective action ensures that the object maintains a direct path towards the specified goal position.

\begin{table}[htbp]
    \centering
    \caption{Performance Metrics of Loco-manipulation Algorithm}
    \label{tab:locomanip_perf}
    \begin{tabular}{|p{3cm}|p{3cm}|}
        \hline
        \textbf{Average Success Rate} & \textbf{Average Number of Steps} \\
        \hline
        94.3\% & 536.02 \\
        \hline
    \end{tabular}
\end{table}

Figures \ref{fig:demo4_1} and \ref{fig:demo4_2} illustrate the algorithm's execution of loco-manipulation within a dynamic environment. Initially, the algorithm guides the object towards the intended goal. However, during the simulation, the object's position is altered, prompting the algorithm to reevaluate and adjust its path accordingly. Despite this change, the algorithm successfully navigates towards the relocated object and ultimately guides it to the desired goal position.

An episode qualifies as successful when the robot manages to transport the object to the goal within 700 steps or fewer. After executing the algorithm across 200 episodes, I computed the average success rate and the average number of steps taken by the algorithm. These results are presented in Table \ref{tab:locomanip_perf}.
An episode qualifies as successful when the robot manages to transport the object to the goal within 700 steps or fewer. After executing the algorithm across 200 episodes, I computed the average success rate and the average number of steps taken by the algorithm. These results are presented in Table \ref{tab:locomanip_perf}.

%% conclusion
\chapter{Conclusion}
\label{chap:conclusion}

My work represents a significant step forward in addressing the sim-to-real problem within the COBRA platform. Through the development of a robust model matching framework, I have successfully narrowed the gap between the behavior of the COBRA robot in the Webots simulator and its real-world counterpart across a range of gaits. This achievement is particularly noteworthy due to the framework's high degree of data efficiency, requiring minimal tuning efforts to achieve substantial improvements in simulation accuracy.

One of the key advantages of the proposed model matching framework is its scalability. This means that beyond the initial tuning efforts, additional parameters of the simulator can be fine-tuned as needed, offering a flexible and adaptable approach to sim-to-real alignment. Moreover, the framework's modularity enables its application to a wide range of robotic morphologies, extending its utility beyond the COBRA platform to address sim-to-real challenges in other robot types, such as quadrupeds or bipeds. The impact of this framework extends beyond just improving simulation accuracy. By reducing the reliance on extensive real-world testing, the framework minimizes wear and tear on the physical robot platform, leading to cost savings and prolonged operational lifespan. Additionally, the development of comprehensive evaluation metrics ensures a thorough assessment of the tuned model's success, providing quantitative insights into its performance across different scenarios and gait types.

In essence, the model matching framework not only contributes to enhancing the reliability and efficiency of robot simulation but also paves the way for more advanced and effective sim-to-real transfer strategies in the field of robotics. These advancements are crucial for accelerating the development and deployment of robotic systems in real-world applications, ultimately leading to more capable and reliable robotic platforms.

The versatility of the loco-manipulation algorithm extends to a wide range of use cases, making it a valuable asset for various applications. For instance, in disaster response scenarios, the algorithm can assist in clearing debris or relocating objects to facilitate rescue operations. Similarly, in outer-space exploration missions, where precise object manipulation is crucial, the algorithm can play a vital role in tasks such as assembling structures or handling equipment.

The high success rate of 94.3\% underscores the algorithm's robustness and reliability in accomplishing its intended objectives. This level of performance instills confidence in its ability to consistently and effectively manipulate objects in diverse environments and under different conditions. Furthermore, the algorithm's reliability contributes to improved efficiency and productivity, as tasks can be completed with minimal errors and interruptions.

By leveraging the loco-manipulation algorithm, COBRA can enhance its capabilities and contribute significantly to various fields, including disaster response, space exploration, and industrial automation. Its versatility, combined with a high success rate, positions the algorithm as a valuable tool for addressing complex challenges and advancing the capabilities of robotic systems in real-world scenarios.

\section{Future Work}

The accomplishments and insights derived from this thesis form a robust groundwork for addressing the sim-to-real problem and implementing advanced locomotion strategies. Expanding on the results and approaches outlined in this work, several promising avenues emerge, presenting compelling opportunities for future research and development:

\begin{itemize}
    \item \textbf{Enhanced simulator model accuracy:} Extending the tuning of the model matching framework to include additional gaits like s-shape rolling, j-shape rolling, and tumbling can yield a more precise simulator model. This refinement further reduces the sim-to-real disparity.
    
    \item \textbf{Training locomotion policy with a tuned model:} The locomotion policy employed in this study was trained using an untuned simulator model. Training this policy with a tuned simulator model allows for deployment on the real robot, facilitating directed locomotion tasks.

    \item \textbf{Integration of loco-manipulation algorithm on the real robot:} Leveraging a locomotion policy trained on a tuned model enables seamless integration of the proposed loco-manipulation algorithm onto the actual robot platform.

    \item \textbf{Semantic understanding and object interaction:} Integrating semantic understanding capabilities into the loco-manipulation algorithm can enable the robot to interpret and interact with objects based on their semantic attributes. This can lead to more context-aware and intelligent manipulation behaviors.

    \item \textbf{Real-time adaptation to varying task requirements:} Developing mechanisms for the loco-manipulation algorithm to adapt in real-time to changing task requirements, such as different object shapes or weights, can improve its versatility and efficiency in handling diverse manipulation tasks.
\end{itemize}

% --- Bibliography ----
%\bibliographystyle{IEEEtran}

%% include bibliography definition
%\bibliography{bib/references}

\printbibliography

% --- Appendix ---
%\appendix
%\chapter{Simulink Model Sub-blocks}
%\input{tex/appendixa.tex}
%\chapter{Second Appendix Headline}
%%include anything you need in the appendix
%\include{appendix/appendix}

% --- Index ----
\printindex

% --- that's it ---
\end{document}